\documentclass{article} %
\usepackage{iclr2026_conference,times}
\iclrfinalcopy

\usepackage{microtype}
\usepackage{hyperref}
\usepackage{url}
\usepackage{booktabs}

\usepackage{xcolor}         %
\usepackage{xspace}         %
\usepackage{colortbl}
\usepackage{color}
\usepackage{microtype}
\usepackage{hyperref}
\usepackage{url}
\usepackage{tcolorbox}
\usepackage{booktabs}
\usepackage{times}
\usepackage{latexsym}
\usepackage{framed}
\usepackage{graphicx}
\usepackage{booktabs}
\usepackage{fancyvrb}
\usepackage{wrapfig}
\usepackage{enumerate}
\usepackage{rotating}
\usepackage{cuted}

\usepackage{soul}
\usepackage{tipa}

\usepackage[utf8]{inputenc} %
\usepackage[T1]{fontenc}    %
\usepackage{hyperref}       %
\usepackage{url}            %
\usepackage{booktabs}       %
\usepackage{amsfonts}       %
\usepackage{nicefrac}       %
\usepackage{microtype}      %
\usepackage{xcolor}         %

\usepackage{amsmath}
\usepackage{multirow}
\usepackage{array}
\usepackage{siunitx}
\usepackage{booktabs}
\usepackage{comment}
\usepackage{tikz}
\usepackage{listings}
\definecolor{LightCyan}{rgb}{0.75,1,1}

\definecolor{pos}{RGB}{167, 199, 231}
\definecolor{neg}{RGB}{250, 160, 160}
\definecolor{amaranth}{rgb}{0.9, 0.17, 0.31}
\definecolor{kellygreen}{rgb}{0.3, 0.73, 0.09}
\definecolor{azure}{rgb}{0.0, 0.5, 1.0}

\usepackage{titlesec}

\titlespacing*{\paragraph}{0pt}{0.5ex plus 0.5ex minus 0.2ex}{1em}

\usepackage{inconsolata}

\newcommand{\modelname}{Ettin}
\newcommand{\modelnamepretty}{\textsc{Ettin}}

\definecolor{darkblue}{rgb}{0, 0, 0.5}
\hypersetup{colorlinks=true, citecolor=darkblue, linkcolor=darkblue, urlcolor=darkblue}

\title{\modelnamepretty: Analyzing Encoders vs Decoders \\ Using the Same Architecture and Data}
\title{Seq vs Seq: An Open Suite \\ of Paired Encoders and Decoders}

\author{
    \textbf{Orion Weller}$^{\hspace{.1em}
    \hspace{.1em}{\color{blue}\boldsymbol{\iota}}}$
    \quad
    \textbf{Kathryn Ricci}$^{\hspace{.1em}\color{blue}\boldsymbol{\iota}}$ 
    \quad
    \textbf{Marc Marone}$^{\hspace{.1em}\color{blue}\boldsymbol{\iota}}$ \\
    \hspace{.03em}
    \textbf{Antoine Chaffin}$^{\hspace{.1em}\color{blue}\boldsymbol{\alpha}}$
    \quad 
    \quad
    \textbf{Dawn Lawrie}$^{\hspace{.1em}\color{blue}\boldsymbol{\iota}}$
    \quad 
    \textbf{Benjamin Van Durme}$^{\hspace{.1em}\color{blue}\boldsymbol{\iota}}$
    \vspace{.5em}\\
    $^{\color{blue}\iota\hspace{.1em}}$Johns Hopkins University
    \quad
    $^{\color{blue}\alpha\hspace{.1em}}$LightOn
     \vspace{.5em}\\
}

\begin{document}

\maketitle

\begin{abstract}
The large language model (LLM) community focuses almost exclusively on decoder-only language models, since they are easier to use for text generation.
However, a large subset of the community still uses encoder-only models for tasks such as classification or retrieval. 
Previous work has attempted to compare these architectures, but is forced to make comparisons with models that have different numbers of parameters, training techniques, and datasets.
We introduce the SOTA open-data \modelnamepretty\footnote{Named for the two-headed mythological Norse giant, symbolizing the two language models heads.} suite of models: paired encoder-only and decoder-only models ranging from 17 million parameters to 1 billion, trained on up to 2 trillion tokens. 
Using the same recipe for both encoder-only and decoder-only models produces SOTA recipes in both categories for their respective sizes, beating ModernBERT as an encoder and Llama 3.2 and SmolLM2 as decoders.
Like previous work, we find that encoder-only models excel at classification and retrieval tasks while decoders excel at generative tasks. 
However, we show that adapting a decoder model to encoder tasks (and vice versa) through continued training is subpar compared to using only the reverse objective (i.e. a 400M encoder outperforms a 1B decoder on MNLI, and vice versa for generative tasks). 
We open-source all artifacts of this study including training data, training order segmented by checkpoint, and 200+ checkpoints to allow future work to analyze or extend all aspects of training.\footnote{Models, code, and data are available at \url{https://github.com/JHU-CLSP/ettin-encoder-vs-decoder}
}
\end{abstract}

\section{Introduction}
The rise of neural language models (LMs) was spurred by encoder-only models such as ELMo \citep{peters2018dissecting} and BERT \citep{bert}. However, the community generally shifted to decoder-only (i.e. GPT-style, ala \citet{gpt3}) models due to their exceptional performance at sequence generation. Due to this lack of popularity for encoder-only models there was limited new model development, thus, we still frequently see usage of older models (i.e. from 2019) by the subset of the community focused on retrieval/classification or fast on-device inference. Although nascent work is attempting to revive encoder-only development \citep{samuel2024berts,warner2024smarter,lee2025clinical}, there still exists a wide gap between the development of encoder-only and decoder-only models (synonymously referred to in this work as \textit{encoders} or \textit{decoders}).

Part of this gap is due to the sentiment within the community that decoders can be adapted for use in tasks that were once predominantly encoder-focused (e.g. classification, embeddings), especially as they can often be used in a zero-shot fashion (i.e. without fine-tuning) \citep{behnamghader2024llm2vec}. As decoder models are more studied, more over-trained \citep{hoffmann2022training}, and are generally larger, they are now claiming the top spots of leaderboards for previously encoder-centric tasks \citep{enevoldsen2025mmteb}. 

Many works have challenged this assumption by comparing encoder-only and decoder-only models of roughly the same sizes \citep{ethayarajh2019contextual,charpentier-samuel-2024-bert,harrag2021bert}. However, these analyses have to be done with incomparable models: using different architectures, different pre-training data, different learning schedules, etc.

Our work aims to provide the foundation to compare encoder-only and decoder-only models by open-sourcing a suite of models trained with the same data, the same architecture, and the same training recipe. Our \modelnamepretty\ suite contains 10 models (5 pairs) ranging from 17 million to 1 billion parameters, and trained for up to 2 trillion tokens. This allows us to quantify the differences between these models (including the effects of scaling parameter size) in an apples-to-apples comparison.

Our models provide state-of-the-art performance for their size among open-data models. Surprisingly, they do so in both encoder settings (w.r.t. ModernBERT) and decoder settings (w.r.t. LLaMA 3.2 and SmolLM2) despite using the same recipe. Notably, our work also provides the first open-data replication of ModernBERT, allowing the community to further build upon our recipe.

We find that, like previous work, encoders excel at classification and retrieval while decoders excel at generative tasks. However, we go beyond previous work to examine the increasingly common setting \citep{behnamghader2024llm2vec} where decoder-models are continued trained as encoders (i.e. cross-objective training). We show results for this cross-objective training in both directions: training encoders for causal language modeling (CLM) and decoders with masked language modeling (MLM).  We find that despite continued training for much longer than previous work (50B tokens) these models do not surpass those that started with this objective, i.e. a 400M encoder outperforms a 1B decoder continue-trained with MLM on MNLI, and vice versa for generative tasks.

Our work also provides the ability to compare these training objectives on other aspects, comparing how they learn. We provide a case study showing the effects of these objectives on gender bias.

Overall, our work provides the first suite of models enabling a fair comparison between encoder-only and decoder-only architectures (while also showing SOTA performance), enabling future work to analyze the effects of these training objectives on downstream tasks.

\section{Related Work}
We describe encoder models as the community is generally more familiar with decoder LMs.\footnote{For those interested in decoders, please see early works such as GPT-2 \citep{radford2019language} and modern models such as OpenAI's GPT-4 \citep{achiam2023gpt}, Google's Gemini \citep{team2023gemini}, Alibaba's Qwen series \cite{yang2025qwen3}, and Meta's LLaMA models \citep{grattafiori2024llama}} It is worth noting that our approach was inspired by Pythia \citep{pythia} which was the first to explore open-data decoder-only models at multiple sizes.

\paragraph{Encoder-only Models}
Encoder-only architectures were the predominant architecture for early transformer models, popularized by models such as BERT \citep{bert}, RoBERTa \citep{roberta}, and DeBERTa \citep{debertav3}. These models showed significantly improved performance over the previous SOTA LSTM models on classification and retrieval tasks. This created a flurry of activity in the encoder space, with models improving on the BERT recipe: the RTD objective from DeBERTa, better data and objectives from RoBERTa, and many smaller variants such as TinyBERT \citep{jiao2019tinybert}, DistilBERT \citep{sanh2019distilbert}, BERT-small \citep{turc2019well}, and MiniLM L12 \citep{wang2020minilm}.
However, these encoders lacked the easy ability to generate text and have generally fallen out of popularity in favor of decoder-only GPT-2 style models. 

Despite this shift, encoders still maintain frequent usage for many tasks that don't require generative output. For example, in March 2025 alone, BERT-base had 90 million downloads on HuggingFace. 

Recently, there has been renewed interest in encoders, as demonstrated by NomicBERT \citep{nomic}, mosiacBERT \citep{mosaic}, and ModernBERT \citep{warner2024smarter}. Unfortunately, ModernBERT (the most performant) does not provide access to their training data. Hence, we use publicly available data sources in order to replicate the training process.

\paragraph{Comparisons between Encoders and Decoders}
Previous work has compared encoder-only and decoder-only models on a wide assortment of tasks. For example \citet{charpentier-samuel-2024-bert} compare DeBERTa and GPT-2 in similar sizes. Other work \citep{yang2023recent,qu2020text,zheng2021adapting,rehana2023evaluation,nielsen2024encoder} compares them on downstream tasks.

However, all of these comparisons have the same underlying limitation: the models they are comparing have different numbers of parameters, different architectures, different training recipes, and different pre-training data. Although some work has attempted to address this \citep{charpentier-samuel-2024-bert,gisserotboukhlef2025pretrainencodersmaskedlanguage}, they have only done so in limited settings with very small amounts of pre-training data. In contrast, we train SOTA models, allowing for an exact comparison.

\paragraph{Bidirectional Attention for Decoders} Although we cannot cover it all here, there have been attempts to use bidirectional attention for standard decoder usage. This includes prefix LM attention \citep{artetxe2022role,chowdhery2023palm,Du2021GLMGL} and other mixed training such as BiTune \citep{kopiczko2024bitune}. However, most modern LM training still favors pure CLM.\footnote{To the best of our knowledge, as much of the details of the best LMs now goes unpublished.}

\paragraph{Checkpoint-Level Model Analyses}
There has also been much work exploring how models learn via their training data. This was popularized by the Pythia \citep{biderman2023pythia} paper and includes many aspects of learning such as data quality and selection \citep{longpre2024pretrainer}, how frequency of entities impacts model learning \citep{oh2024frequency}, effects of the recency of data \citep{cheng2024dated}, and whether you can recognize and extract training data from models \citep{zhang2024min}. Our work allows these experiments to be done on more recent SOTA models and provides a way to compare encoders and decoders on various facets of learning.

\begin{table}[t!]
\small
\centering
\begin{tabular}{lcccccc}
\toprule
 & \textbf{17M} & \textbf{32M} & \textbf{68M} & \textbf{150M} & \textbf{400M} & \textbf{1B} \\
\textbf{Parameter} & \textbf{(XXS)} & \textbf{(XS)} & \textbf{(Small)} & \textbf{(Base)} & \textbf{(Large)} & \textbf{(XL)} \\
\midrule
Layers & 7 & 10 & 19 & 22 & 28 & 28 \\
Hidden Size & 256 & 384 & 512 & 768  & 1024 & 1792 \\
Intermediate Size & 384 & 576 & 768 & 1152 & 2624 & 3840 \\
Attention Heads & 4 & 6 & 8 & 12 & 16 & 28 \\
Learning Rate & 3e-3 & 3e-3 & 3e-3 & 8e-4 & 5e-4 & 5e-4 \\
Weight Decay & 3e-4 & 3e-4 & 3e-4 & 1e-5 & 1e-5 & 5e-5 \\
Warmup Tokens (B) & 4 & 4 & 3 & 3  & 2 & 2 \\
BS Warmup (B) & 125 & 100 & 75 & 50 & 10 & 3 \\
\bottomrule
\end{tabular}
\vspace{0.75em}
\caption{Configuration for each \modelname\ model size. Both encoders and decoders use the exact same configuration, differing only in attention (bidirectional vs causal) and objective (MLM vs CLM).}
\vspace{-1em}
\label{tab:params}
\end{table}

\section{Experimental Settings}
\subsection{Training Data}
We create an open-source replication of ModernBERT \citep{warner2024smarter} due to it being the strongest publicly available encoder-only model and using comparable techniques to decoder-only models. This provides the best starting place for a recipe that spans both training objectives. However, ModernBERT's data is not publicly available -- thus, we aim to replicate the recipe using open-data. 

We do so by pulling from the best publicly available datasets used for training decoder-only models, such as Olmo \citep{olmo,olmo2}. Thus, we use a mix of DCLM \citep{li2024datacomp} combined with various curated sources from Dolma v1.7 \citep{soldaini2024dolma}. In the process of training, the Olmo 2 paper \citep{olmo2} described their approach of using filtered DCLM and other higher-quality sources for the decay phase (similar to FineWeb-Edu filtering \citet{penedo2024fineweb,lozhkov2024finewebedu}). We decided to use these newer sources for our later phases.\footnote{We ablated with the non-filtered data and found worse results.}

To allow others to easily extend our work, we provide both formats: the data which can be used for training as well as the data seen by the models in batch order for future analyses.

\subsection{Architecture}
As ModernBERT has only two sizes, we develop new shapes for our smaller and larger models (Table~\ref{tab:params}). We aim to follow the design espoused by MobileLLM \citep{liu2024mobilellm} with deep but thin models. However, for the 1B model, we keep the same number of layers but make the model wider. We choose models parameter sizes at roughly 2x increments while matching common encoder sizes, e.g. 17M, 32M, 68M, 150M, 400M, and 1B. For a detailed list of the differences, see Table~\ref{tab:params}.

\subsection{Training Recipe}
We use the same general process described by open-data models (which was followed by ModernBERT) for training both encoder-only and decoder-only models -- with a few specific changes for the encoder architecture (i.e. masking ratio). In summary, we include three general phases: base pre-training, mid-training/context extension, and decay. See Table~\ref{tab:combined_training_mix} for the precise sources of training data in each phase. We use a trapezoidal learning rate scheduler, with general hyperparameters shown in Appendix~\ref{app:arch} and size-dependent hyperparameters in Table~\ref{tab:params}.  For compute details see Appendix~\ref{app:compute}.

\begin{table}[t!]
  \centering
  \small
\begin{tabular}{@{}l@{\hspace{4pt}}l@{\hspace{4pt}}r@{\hspace{2pt}}r@{\hspace{4pt}}r@{\hspace{2pt}}r@{\hspace{4pt}}r@{\hspace{2pt}}r@{}}
    \toprule
    & & \multicolumn{2}{c}{\textbf{Pre-training}} & \multicolumn{2}{c}{\textbf{Mid-training}} & \multicolumn{2}{c}{\textbf{Decay Phase}} \\
    \cmidrule(lr){3-4} \cmidrule(lr){5-6} \cmidrule(lr){7-8}
    \textbf{Category} & \textbf{Dataset} & \textbf{Tokens (B)} & \textbf{\%} & \textbf{Tokens (B)} & \textbf{\%} & \textbf{Tokens (B)} & \textbf{\%} \\
    \midrule
    News & CC News & 7.3 & 0.4 & -- & -- & -- & -- \\
    Code & Starcoder & 263.9 & 15.5 & 38.4 & 15.4 & -- & -- \\
    Code & Code\_Repos & -- & -- & -- & -- & 20.2 & 26.5 \\
    Crawl & CC Head & 356.6 & 20.9 & -- & -- & -- & -- \\
    Crawl & DCLM & 837.2 & 49.1 & -- & -- & -- & -- \\
    Crawl & DCLM (Dolmino) & -- & -- & 175.5 & 70.4 & 26.0 & 34.1 \\
    Math & Open-Web-Math & 12.7 & 0.7 & -- & -- & -- & -- \\
    Math & Algebraic StackExchange & 12.6 & 0.7 & -- & -- & -- & -- \\
    Math & Math (Dolmino) & -- & -- & 10.4 & 4.2 & 5.0 & 6.6 \\
    Scientific & PeS2o & 57.3 & 3.4 & 8.3 & 3.3 & -- & -- \\
    Scientific & Arxiv & 28.0 & 1.6 & 4.1 & 1.6 & 3.0 & 3.9 \\
    Social & Reddit & 80.3 & 4.7 & 6.2 & 2.5 & -- & -- \\
    Social & StackExchange & 19.6 & 1.1 & -- & --& -- & -- \\
    Social & StackExchange (Dolmino) & -- & -- & 2.7 & 1.1 & 4.0 & 5.2 \\
    Reference & Textbooks & -- & -- & -- & -- & 0.5 & 0.7 \\
    Reference & Dolma Books & 5.3 & 0.3 & 0.8 & 0.3 & 10.5 & 13.8 \\
    Reference & Wikipedia & 7.3 & 0.4 & 0.5 & 0.2 & 3.0 & 3.9 \\
    Instruction & Tulu Flan & 16.6 & 1.0 & 2.4 & 1.0 & 4.1 & 5.4 \\
    \midrule
    \textbf{Total} & & 1,704.7 & 100.0 & 249.3 & 100.0 & 76.3 & 100.0 \\
    \bottomrule
  \end{tabular}
  \vspace{0.75em}
    \caption{Training data mixture across the various training stages (pre-training, mid-training, decay). Later stages use higher quality data, from the recently released Dolmino dataset \citep{olmo2}. Dashes indicate that no data from that source was used. We trained for 1.7T tokens for pre-training, 250B for mid-training, and 50B for the decay phase. We sample from the dataset and repeat (or under-sample) as needed to hit the token counts used for training.}
    \vspace{-1em}
  \label{tab:combined_training_mix}
\end{table}

The only differences between the encoder and decoder models are: (1) the objective function, i.e. masked language modeling (MLM) for the encoder\footnote{For the encoder we use a 30\% masking ratio except for the decay phase, which is 15\%.} vs causal language modeling (CLM) for the decoder and (2) the attention pattern, i.e. causal for the decoder and bidirectional for the encoder.

We checkpoint every 8.5B tokens, with 236 checkpoints per model. Combined with the batch ordering of the data, this enables precise pinpointing of what the model learned between each checkpoint.

\paragraph{Base Pre-training} This stage encompasses the warmup and stable phase of the trapezoidal learning rate, training for 1.7T tokens. We use both learning rate and batch size warmup. The data in this stage comprises a wide mix of sources to allow for general learning.

\paragraph{Context Extension / Mid-Training} In this phase we increase the quality of the data and change both the data and base RoPE \citep{su2024roformer} to handle longer context. We update the data length to be up to 8000 tokens and RoPE parameters to 160k (for global and local layers). For the data, we drop the noisiest sections (older Dolma common crawl, CC News, general StackExchange) and include  filtered DCLM, math, and StackExchange. We then train for 250B tokens and use an inverse square root learning rate schedule from the peak learning rate to 1/2 of the peak. 

\paragraph{Decay Phase} Finally, we use one more inverse square root learning rate schedule to decay for 50B tokens. We follow the general ProLong recipe \citep{gao2024train}, increasing long context data such as Dolma books, Wikipedia, and open-access textbooks. We decay to 0.02 of the peak LR.

\subsection{Major training differences from ModernBERT} A concise summary of the largest differences from the ModernBERT recipe are (1) the use of open-data, (2) decay in the context extension phase, (3) no model merging,\footnote{We do this for ease of scientific comparison, however, if one was to use this for downstream applications a simple merge would likely boost performance another point or two.} (4) a lower masking ratio for the decay phase (15\% instead of 30\%) and (5) local and global RoPE to be the same value.

\subsection{Cross-Objective Training}
As encoder models have gone out of popularity, decoder models have increased in size (both parameters and pre-training data). Thus, these newer decoder models are typically trained for much longer than previous encoder models (i.e. BERT). Due to this it has become common to adapt these larger decoder models to what were previously encoder-centric tasks \citep{qwen3embedding}. With paired encoder and decoder models, we can now answer the question of \textbf{how effective this continued pre-training approach is} and \textbf{whether it is still worth training both types of models}. We call this \textit{cross-objective training}: taking the final model and continue pre-training it on the reverse objective. Following \citet{behnamghader2024llm2vec}, we do not use MLM but rather use MNTP, that is, the masked token is predicted using the hidden state of the previous token to better align with CLM.

We train for 50B tokens with the reverse objective, which is far more than previous work has attempted e.g. LLM2Vec \citep{behnamghader2024llm2vec}, which is around 10B tokens.  Although the ratio of pre-training and cross-objective training is unbalanced, this mimics the realistic setting where the adaptation is done with very small amounts of data comparatively. We do this cross-objective training on the highest quality data we have available, which was used in the last decay phase.\footnote{We note that this means it repeating this data twice, however, as shown by previous work \citep{muennighoff2023scaling} two repetitions on high quality data has no adverse effects.} We use a new trapezoidal learning rate schedule with 3B tokens of warmup and 10B tokens of decay.\footnote{For the 1B model this is scaled by 1/3 again due to compute availability.} Thus, by the end we have an encoder-from-decoder (i.e. a decoder further pretrained with MNTP similar to LLM2Vec\footnote{We use a 15\% masking rate for the encoder-from-decoder as to maintain a middle ground masking ratio.}) and a decoder-from-encoder (i.e. an encoder further pre-trained with CLM).

\section{Experiments}
We aim to compare encoder and decoder models. However, first, to give those experiments credence, we show that our models are SOTA. This strengthens our claim and helps alleviate concerns that we made training choices that favored one architecture over the other -- instead we have SOTA models in both architectures for their sizes, showing our method's effectiveness. Note though, that the purpose of our paper was not to be SOTA overall (i.e. compared to OpenAI, etc.), but to provide a comparison for encoders and decoders. For space and to avoid repetition, specific model size details are in Table~\ref{tab:params}.

\begin{table*}[t!]
\resizebox{\textwidth}{!}{
\centering
\begin{tabular}{l|ccccc|ccc}
\toprule
& \multicolumn{5}{c|}{\textbf{Embedding Tasks}} & \multicolumn{3}{c}{\textbf{GLUE Tasks}} \\
\textbf{Model Name} & \textbf{CSN} & \textbf{MLDR} & \textbf{Clustering} & \textbf{Retrieval} & \textbf{MTEB v2} & \textbf{SST-2} & \textbf{MNLI} & \textbf{GLUE Avg} \\
\midrule
\multicolumn{9}{c}{\textbf{XXS Models (7-17M parameters)}} \\
\midrule
BERT-mini & 41.3 & 16.8 & 39.0 & 34.7 & 49.2 & 88.3 & 77.2 & 76.4 \\
TinyBERT & 39.8 & 14.2 & 37.4 & 33.3 & \textbf{49.7} & \textbf{91.2} & \textbf{80.9} & 77.0 \\
\modelname-Enc-17m & \textbf{59.1} & \textbf{24.4} & \textbf{39.1} & \textbf{35.6} & 48.9 & \textbf{91.2} & 79.5 & \textbf{79.2} \\
\midrule
\multicolumn{9}{c}{\textbf{XS Models (28-33M parameters)}} \\
\midrule
BERT-small & 46.0 & 19.9 & \textbf{39.6} & 38.1 & 51.1 & 90.1 & 79.2 & 79.0 \\
MiniLM L12 & 48.3 & 19.6 & 37.8 & 38.4 & \textbf{51.3} & \textbf{93.3} & \textbf{85.6} & \textbf{84.6} \\
\modelname-Enc-32m & \textbf{69.2} & \textbf{28.4} & \textbf{39.6} & \textbf{39.7} & 50.9 & 92.0 & 83.4 & 83.5 \\
\midrule
\multicolumn{9}{c}{\textbf{Small Models (68-82M parameters)}} \\
\midrule
DistilBERT & 47.9 & 23.7 & 39.8 & 40.8 & \textbf{52.7} & 92.2 & 82.7 & 81.5 \\
DistilRoBERTa & 60.3 & 19.7 & 39.3 & 40.0 & 51.8 & 93.1 & 84.7 & 83.8 \\
\modelname-Enc-68m & \textbf{75.1} & \textbf{30.1} & \textbf{40.1} & \textbf{43.1} & 52.6 & \textbf{94.4} & \textbf{87.0} & \textbf{87.2} \\
\midrule
\multicolumn{9}{c}{\textbf{Base Models (123-150M parameters)}} \\
\midrule
BERT base & 51.0 & 24.8 & 40.4 & 41.2 & 52.9 & 93.1 & 85.4 & 84.7 \\
ModernBERT base & 75.9 & 30.4 & 41.3 & 43.9 & \textbf{54.0} & \textbf{96.0} & 89.1 & 88.4 \\
\modelname-Enc-150m & \textbf{76.3} & \textbf{31.8} & \textbf{41.5} & \textbf{45.7} & \textbf{54.0} & 95.8 & \textbf{89.2} & \textbf{88.9} \\
\midrule
\multicolumn{9}{c}{\textbf{Large Models (353-395M parameters)}} \\
\midrule
BERT large & 54.4 & 25.3 & 41.5 & 42.9 & 53.8 & 93.3 & 86.3 & 85.2 \\
ModernBERT large & 78.3 & 34.9 & 41.5 & 47.0 & 55.0 & \textbf{97.1} & 90.8 & 90.4 \\
\modelname-Enc-400m & \textbf{80.7} & \textbf{36.2} & \textbf{41.8} & \textbf{48.4} & \textbf{55.5} & 96.7 & \textbf{91.3} & \textbf{90.8} \\
\midrule
\multicolumn{9}{c}{\textbf{XL Models (884M-1.2B parameters)}} \\
\midrule
DeBERTa-v1-xl & 75.6 & 28.1 & \textbf{42.5} & 47.2 & \textbf{56.4} & \textbf{97.1} & 91.7 & 90.7 \\
\modelname-Enc-1B & \textbf{82.3} & \textbf{40.2} & 41.9 & \textbf{50.1} & 56.0 & \textbf{97.1} & \textbf{91.8} & \textbf{91.6} \\
\bottomrule
\end{tabular}
}
\caption{\modelnamepretty\ encoders compared to other encoder-only models across various sizes on retrieval and GLUE tasks. Due to space, we show two representative tasks from MTEB v2 and two from GLUE, as well as a code-based retrieval evaluation (CodeSearchNet) and a long-context evaluation (MLDR). See Appendix~\ref{app:all_encoder} for the full tables of GLUE and MTEB v2. \textbf{\modelnamepretty\ shows significant gains over baseline encoders, including ModernBERT, while also having both larger and smaller sizes.}}
\label{tab:combined_results}
\vspace{-1em}
\end{table*}

\subsection{Individual Evaluations}

\paragraph{Encoder-Only Results}
We use two baselines for each size type: extra extra small (XXS) BERT-mini and TinyBERT, extra small (XS) models MiniLM L12 and BERT-small, small (S) models DistilBERT and DistilRoBERTa, base (B) models BERT and ModernBERT, large (L) models BERT-large and ModernBERT-large, and an extra large (XL) model DeBERTA v2 XL.\footnote{We also ran experiments with DeBERTa XXL as shown in the appendix. However, due to the size and slowness of the architecture we could not do a comparable grid search. Our results in the appendix are after 300 days of H100s hours, but still did not complete the full sweep. Furthermore, as DeBERTa XXL is > 1.5B we exclude it as it is significantly larger than 1B (i.e. > 50\% larger).}

We evaluate on various encoder tasks, including GLUE \citep{wang2018glue}, MTEB v2 English \citep{enevoldsen2025mmteb}, MDLR for long context \citep{bge-m3}, and CodeSearchNet for code evaluation \citep{husain2019codesearchnet}. We use the same evaluation setup as ModernBERT for the evaluation for an equal comparison (see Appendix~\ref{app:encoder_sweep} for hyperparameter details).

We find in Table~\ref{tab:combined_results} that \modelnamepretty\ compares favorably overall. The relatively larger gains in the bigger sizes is likely because the smaller model baselines are heavily optimized with distillation.\footnote{We also note that MiniLM L12 has twice the amount of non-embedding parameters 21M vs 12M.} Even so, we see that they generally outperform the baselines without doing any distillation: e.g. \modelname-68m with a GLUE average of 87.2 compared to the next best DistilRoBERTa at 83.8. Even for the more recent ModernBERT baselines we see improved performance (88.9 GLUE average vs 88.4 for the base size). Thus we can see that \modelnamepretty\ matches or improves the SOTA for encoder-only models.

\begin{table*}[t!]
\resizebox{\textwidth}{!}{
\centering
\begin{tabular}{l|cccccccccc|c}
\toprule
\textbf{Model Name} & \textbf{ARC} & \textbf{HS} & \textbf{LMB} & \textbf{OBQA} & \textbf{PIQA} & \textbf{SciQ} & \textbf{SIQA} & \textbf{TQA} & \textbf{WG} & \textbf{WSC} & \textbf{Avg} \\
\midrule
\multicolumn{12}{c}{\textbf{XXS Models (14-17M parameters)}} \\
\midrule
Pythia-14m  & 21.2 & 26.0 & 7.1 & 26.2 & 55.2 & 43.8 & 33.4 & 0.0 & 50.3 & \textbf{51.6} & 31.5 \\
\modelname-Dec-17m & \textbf{21.3} & \textbf{27.1} & \textbf{23.0} & \textbf{27.2} & \textbf{57.7} & \textbf{71.1} & \textbf{35.4} & \textbf{2.6} & \textbf{50.9} & 48.0 & \textbf{36.4} \\
\midrule
\multicolumn{12}{c}{\textbf{XS Models (32M parameters)}} \\
\midrule
\modelname-Dec-32m  & \textbf{23.5} & \textbf{28.5} & \textbf{28.5} & \textbf{28.2} & \textbf{57.7} & \textbf{77.5} & \textbf{36.4} & \textbf{3.8} & \textbf{53.1} & \textbf{50.2} & \textbf{38.7} \\
\midrule
\multicolumn{12}{c}{\textbf{Small Models (68-82M parameters)}} \\
\midrule
DistilGPT & 23.0 & 27.5 & 25.0 & 26.8 & 59.8 & 62.6 & 36.1 & 0.3 & \textbf{50.4} & 53.8 & 36.5 \\
\modelname-Dec-68m  & \textbf{25.3} & \textbf{33.4} & \textbf{35.2} & \textbf{29.4} & \textbf{61.8} & \textbf{83.2} & \textbf{38.8} & \textbf{5.6} & 50.1 & \textbf{55.3} & \textbf{41.8} \\
\midrule
\multicolumn{12}{c}{\textbf{Base Models (135-160M parameters)}} \\
\midrule
SmolLM2-135m  & \textbf{29.1} & \textbf{43.1} & 42.9 & \textbf{32.4} & \textbf{68.4} & 78.5 & 39.4 & 5.0 & \textbf{53.7} & \textbf{59.7} & 45.2 \\
Pythia-160m & 24.0 & 30.2 & 32.9 & 26.4 & 62.0 & 67.2 & 36.9 & 0.4 & 52.4 & 58.2 & 39.1 \\
\modelname-Dec-150m  & 28.6 & 40.3 & \textbf{43.2} & 29.2 & 66.6 & \textbf{89.6} & \textbf{40.1} & \textbf{11.2} & \textbf{53.7} & 59.0 & \textbf{46.2} \\
\midrule
\multicolumn{12}{c}{\textbf{Large Models (360-410M parameters)}} \\
\midrule
Pythia-410m  & 24.7 & 40.6 & 51.5 & 29.4 & 67.0 & 72.3 & 39.0 & 1.8 & 53.6 & 65.2 & 44.5 \\
SmolLM2-360m  & \textbf{37.6} & \textbf{56.3} & \textbf{53.5} & \textbf{37.6} & \textbf{71.8} & 86.6 & 40.7 & \textbf{18.4} & \textbf{58.6} & 70.3 & \textbf{53.1} \\
\modelname-Dec-400m & 33.6 & 54.3 & 52.3 & 34.4 & 71.0 & \textbf{91.8} & \textbf{45.5} & 18.3 & 57.6 & \textbf{71.8} & \textbf{53.1} \\
\midrule
\multicolumn{12}{c}{\textbf{XL Models (908M-1.2B parameters)}} \\
\midrule
OLMo-1B-0724  & 32.3 & \textbf{66.1} & 61.0 & 35.6 & \textbf{75.1} & 91.8 & \textbf{49.2} & 1.2 & 61.6 & 76.9 & 55.1 \\
Llama-3.2-1B  & 36.2 & 63.7 & \textbf{62.1} & 37.2 & 75.0 & 88.4 & 43.2 & 24.9 & 60.6 & 74.7 & 56.6 \\
\modelname-Dec-1B  & \textbf{39.7} & 62.9 & 58.4 & \textbf{41.6} & 74.4 & \textbf{93.8} & 48.2 & \textbf{29.3} & \textbf{62.7} & \textbf{79.1} & \textbf{59.0} \\
\bottomrule
\end{tabular}
}
\caption{Performance comparison of \textbf{decoder-only} models across tasks, organized by size categories. \textbf{We see that \modelnamepretty\ decoders compare favorably, matching or exceeding the previous open-data SOTA}. Task names in order are ARC, Hellaswag, LAMBADA, OpenBookQA, Social IQA, TriviaQA, Winogrande, and Winograd Schema Challenge.}
\vspace{-1em}
\label{tab:decoder_results}
\end{table*}

\paragraph{Decoder-Only Results}
We use two baselines for each size type when available, but few very small decoder-only LMs exist: One extra extra small (XXS) model Pythia-14M,\footnote{This is likely just a debug-sized run and not an official size, as they do no include it in their paper. However, as not other decoder models in this size could be found, we use it as a reference.} no models in the extra small (XS) category that we could find, one small (S) model DistilGPT \citep{sanh2019distilbert}, two base (B) models Pythia 160M  \citep{biderman2023pythia} and SmolLM2 135M, two large (L) models Pythia 410M and SmoLM2 360M \citep{allal2025smollm2}, and extra large (XL) models Olmo 1B 0724 \cite{olmo} and Llama 3.2 1B \citep{llama3}.\footnote{What is considered ``1B" has a large range, up to nearly 2B parameters. We thus restrict our range to < 1.2B parameters to be a ``1B" model, which excludes models like SmolLM2 1.7B and Olmo 2 1B (actually 1.5B).}

We evaluate on a wide range of tasks using the Eleuther AI harness \citep{eval-harness} (see Appendix~\ref{app:eval} for details), consolidating tasks used in the Pythia and SmolLM papers including: the ARC Challenge (ARC) \cite{clark2018think}, HellaSwag (HS) \citep{zellers2019hellaswag}, LAMBADA (LMB) \citep{paperno2016lambada}, OpenBookQA (OBQA) \citep{mihaylov2018can}, Social IQA (SIQA) \citep{sap2019socialiqa}, TriviaQA (TQA) \citep{joshi2017triviaqa}, Winogrande (WG) \citep{sakaguchi2021winogrande}, and the Winograd Schema Challenge (WSC) \citep{ea01b9c0db064caca6986b925d75f2bb}. Note all tasks use zero-shot closed-book evaluations.

We see the results in Table~\ref{tab:decoder_results} and see that \modelnamepretty\ performs well compared to baseline models, such as \modelnamepretty-150m outperforming SmolLM2 46.2 to 45.2 and \modelnamepretty-1B's 59.0 to Llama 3.2 1B's 56.6 average). Thus we can see that \modelnamepretty\ improves the SOTA for open-data decoder-only models.

\begin{figure*}[t!]
  \centering
\includegraphics[width=\textwidth, trim=0 2em 0 1em]{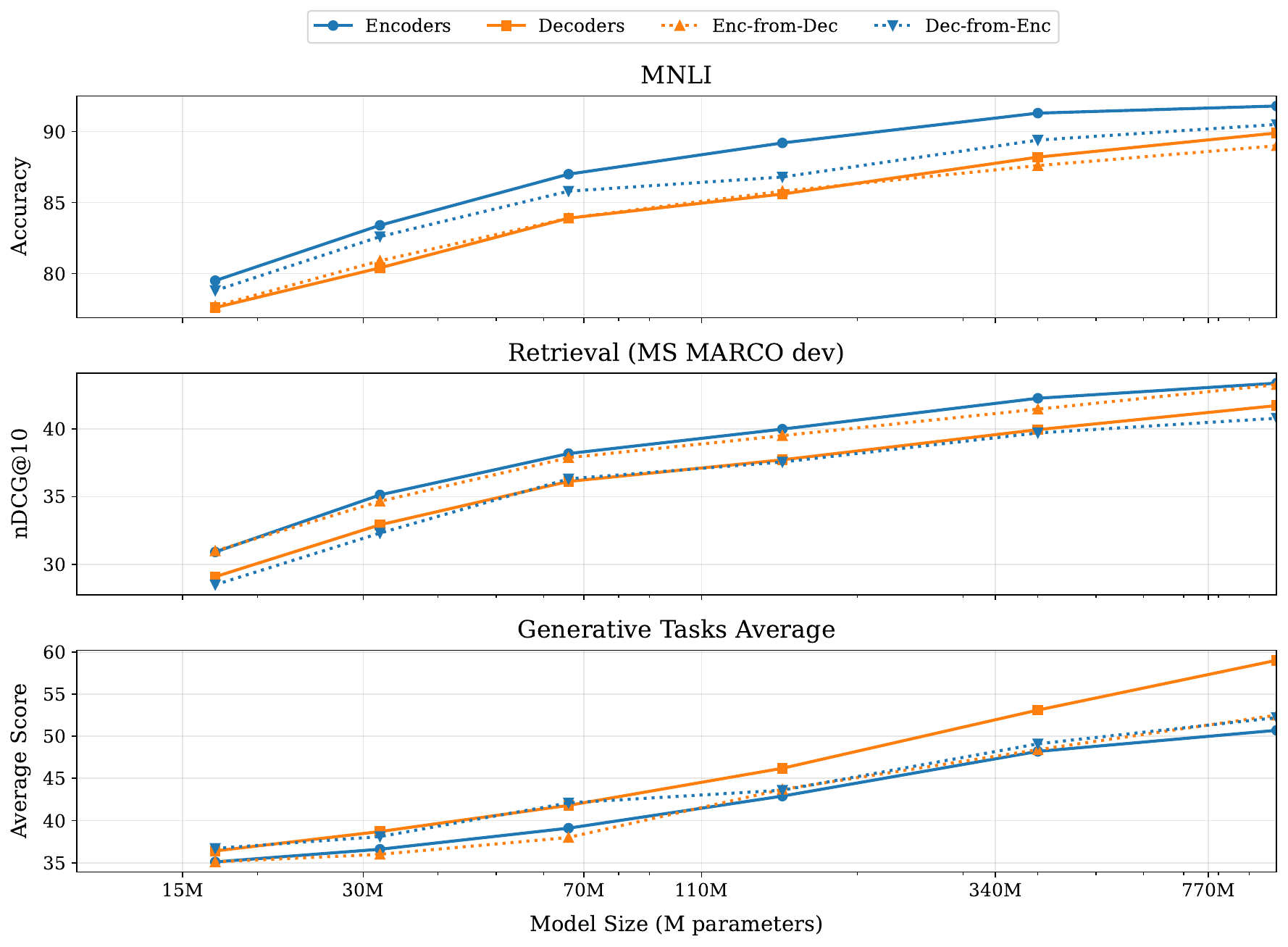}
  \caption{Encoder vs decoder comparison across model size. Generally \textbf{models in the preferred architecture (e.g. encoders in MNLI) do better than the opposite architecture even with an order of magnitude greater size}, e.g. a 400M decoder outperforming the 1B encoder. Notably, in generative tasks, decoders-from-encoders scale poorly with size.
  }
  \label{fig:enc_v_dec}
\vspace{-1em}
\end{figure*}

\subsection{Encoders vs Decoders}
Now that the strength of the training recipe is established, we can compare the two training objectives.

For simplicity, we show the two most representative encoder tasks (MS MARCO dev \citet{msmarco} for retrieval and classification on MNLI) and keep the average generative score.\footnote{As decoder evaluations are significantly quicker than fine-tuning which is required for encoder tasks.}
We evaluate the decoders on encoder-only tasks and vice versa, and similarly with the cross-objective trained models. We evaluate encoder-only models on decoder generative tasks using the method proposed in \citet{samuel2024berts}, i.e. using three mask tokens at the end of the sequence and filling in the first token iteratively.
Figure~\ref{fig:enc_v_dec} shows the results of this comparison across models sizes.

\paragraph{MNLI Classification} On the representative classification task, we see that encoders dominate decoders, as typically found. Furthermore, we see that even cross-objective continued pre-training  does not resolve this gap, with enc-from-dec performance remaining similar to the original decoder model. Furthermore, the pure encoder models are typically better than the next larger sizes of decoders, e.g. the 150M encoder scoring 89.2 compared to the 400M decoder's 88.2.

\paragraph{MS MARCO Dev Retrieval}
For retrieval we see similar encoder dominance like classification, but notably improved performance when continue pre-training the decoder (i.e. the encoder-from-decoder). The MTNP continue pre-training significantly helps the decoder at all sizes, yet even the additional 50B tokens of pre-training is not enough to match the performance of the encoder (i.e. for the 400M size we have 42.2 for the encoder vs 41.4 for the encoder-from-decoder). Although the difference is not as pronounced as in classification, we find that continued pre-training a decoder for retrieval is still subpar compared to simply using an encoder, even despite the additional 50B tokens.

\paragraph{Generative Tasks} We find the reverse of the previous tasks: decoders do better than the decoder-from-encoder in general, with a widening gap (from similar scores at 68m parameters to greater than a 6 point difference at 1B) as model size increases. Notably, it appears that continued training of encoders-from-decoders scales poorly, perhaps why there is little-to-no previous work on the topic. 

Despite this, we note that this average hides some nuance: on ``generative" tasks that are more classification focused (such as ARC and SciQ) encoder models used in a generative fashion actually exceed decoder performance (i.e. for the 400M size the encoder scores 35.6 ARC vs the decoder's 33.6). However, decoders show huge gains on tasks such as HellaSwag, TriviaQA, and SiQA, making it so the average is strongly in favor of the decoders. See Table~\ref{tab:all_generative_results} for all sub-task results. 

Although most of our models are too small to have signal on more difficult, but standard LM tasks like MMLU and GSM8k, we evaluate these for just the 1B sized models in Table~\ref{tab:additional-benchmarks}. We find similar results to the previous table, illustrating that the Decoder-From-Encoder significantly outperforms on MMLU classification but is significantly worse on the generative GSM8k.

\begin{table}[htb!]
\centering
\begin{tabular}{lcc}
\toprule
\textbf{Model} & \textbf{MMLU CS} & \textbf{GSM8k} \\
\midrule
Decoder       & 27.0 & \textbf{32.0} \\
Dec-From-Enc  & \textbf{37.0} & 18.9 \\
\bottomrule
\end{tabular}
\caption{Harder benchmark results for 1B models on reasoning and classification tasks.\label{tab:additional-benchmarks}}
\vspace{-1em}
\end{table}

\section{Case study: Gender Bias}
Due to the \modelname\ suite's open-pretraining data, we can also analyze other aspects of learning across pre-training objectives. As one example, we analyze gender representations for bias. 

We use the WinoGender benchmark \citep{rudinger2018gender} using the ``Gotcha" split that has a 50/50 split of male/female stereotypical pronouns (i.e. female for nurse). However, the standard coreference task is hard for most of our small models. Thus, we show results for an easier task: simply predicting the pronoun in the sentence. For the standard coreference task results, see Table~\ref{tab:winogender_acc} in the appendix.

\begin{figure*}[t!]
  \centering
\includegraphics[width=\textwidth, trim=0 2em 0 1em]{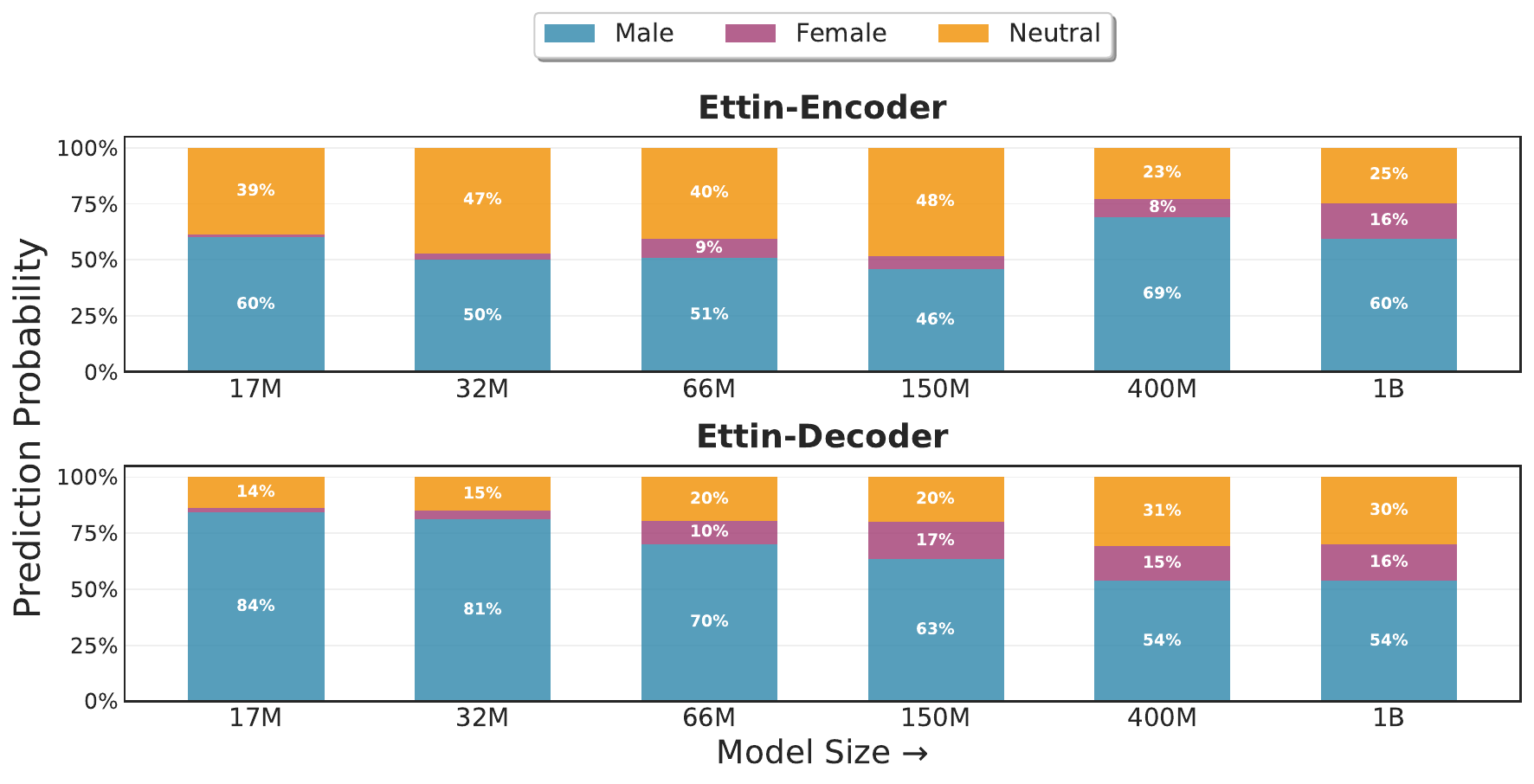}
  \caption{Gender pronoun predictions on the Gotcha split of WinoGender \citep{rudinger2018gender}, a 50/50 stereotypical male/female split. \textbf{We see that encoder models are more likely to use gender neutral pronouns whereas both are biased towards male pronouns.}}
  \label{fig:bias}
 \vspace{-1em}
\end{figure*}

We have each model predict the pronouns (i.e. by using a mask token for encoders or by choosing the lower perplexity sentence with decoders) and show the distribution of predicted pronouns per model (male, female, or gender neutral).\footnote{There are more than three types of pronouns used in English beyond what is in this dataset. However, WinoGender is only designed for these three. We leave extensions of this dataset to future work.}  The results are in Figure~\ref{fig:bias}, which shows that encoders are much more likely to use a gender neutral pronoun overall. In both encoders and decoder, female pronouns become more used as the size of the model gets larger: for decoders there is a clear trend of progressively smaller amounts of male pronouns, whereas for encoders the trend is more stochastic. For effects of the cross-training objectives on the model, see Figure~\ref{fig:bias_all} in the Appendix.

\paragraph{Overall} As both models had the same training data, we find that the MLM objective leads the model to choose more neutral pronouns over female pronouns. However, male gender bias seems strong in both models, if slightly higher for decoders. Thus, this is one example of the analysis enabled by our data; we leave others to future work.

\section{Discussion}
Our work suggests the following conclusions: (1) as speculated by previous work, MLM and CLM objectives do convey different strengths -- MLM for classification and retrieval and CLM for generative tasks. However, (2) we also went a step further to show that simply continued pre-training on the reverse objective does not make up the difference from not using the preferred architecture. 

This has several implications for those using models for classification or retrieval: currently the top models on leaderboards like MTEB are 7B+. However, based on our experiments, it is likely a 3B encoder model would outperform it. But, the lack of large encoder-only models means that approaches that continue pre-train decoders using MLM will likely outperform all other options (as is currently seen on the leaderboards). In the small scale regime (1B or less) where it is easier to train more ``niche" encoder models for classification/retrieval, our results indicate that encoders will continue to outperform all others in their size range (and even ones above it).

Our results also suggest that encoders and decoders learn differently in other aspects as well, such as gender bias. Although this is just one example, we look forward to future research that discovers other differences. Overall, our artifacts allow for a range of new analyses and pre-training research.

We note that concurrent work \citep{gisserotboukhlef2025pretrainencodersmaskedlanguage} did a similar analysis and found that starting from CLM and doing continued MLM pre-training was better in nearly all cases. However, pre-training was done with only 100B tokens; thus it is likely that this is an artifact of CLM's comparative data efficiency in the smaller data regime (i.e. loss on every token compared to loss on X\% of tokens for MLM). Our work shows that at scale, for SOTA models, encoders do greatly outperform encoders-from-decoders (as also shown by SOTA concurrent work \citet{marone2025mmbert}).

\section{Conclusion}
We provide the first suite of paired models that use the same training recipe to compare encoder-only and decoder-only models. Our models are SOTA in their size for open-data models, and are the first public recipe for ModernBERT-style models. We show that encoders are strong in classification and retrieval, while decoders are strong in generative tasks. 
Furthermore, we show that this difference can not easily be solved by continued training with the reverse objective.
We show that this suite allows the analysis of how pre-training objective impacts learning, showing a case study in gender bias.
We release all artifacts (including training data order) to help future researchers analyze these models.

\section*{Acknowledgments}
This work has been supported by both DARPA SciFy and the U.S. National Science Foundation under grant 2204926. Any opinions, findings, and conclusions or recommendations expressed in this article are those of the authors and do not necessarily reflect the views of the National Science Foundation or DARPA. OW is supported by an NSF GRFP fellowship.

\bibliography{colm2025_conference}

@article{peters2018dissecting,
  title={Dissecting contextual word embeddings: Architecture and representation},
  author={Peters, Matthew E and Neumann, Mark and Zettlemoyer, Luke and Yih, Wen-tau},
  journal={arXiv preprint arXiv:1808.08949},
  year={2018}
}

@inproceedings{bert,
  author       = {Jacob Devlin and
                  Ming{-}Wei Chang and
                  Kenton Lee and
                  Kristina Toutanova},
  editor       = {Jill Burstein and
                  Christy Doran and
                  Thamar Solorio},
  title        = {{BERT:} Pre-training of Deep Bidirectional Transformers for Language
                  Understanding},
  booktitle    = {Proceedings of the 2019 Conference of the North American Chapter of
                  the Association for Computational Linguistics: Human Language Technologies,
                  {NAACL-HLT} 2019, Minneapolis, MN, USA, June 2-7, 2019, Volume 1 (Long
                  and Short Papers)},
  pages        = {4171--4186},
  publisher    = {Association for Computational Linguistics},
  year         = {2019},
  url          = {https://doi.org/10.18653/v1/n19-1423},
  doi          = {10.18653/V1/N19-1423},
  bibsource    = {dblp computer science bibliography, https://dblp.org}
}

@inproceedings{debertav3,
  author       = {Pengcheng He and
                  Jianfeng Gao and
                  Weizhu Chen},
  title        = {DeBERTaV3: Improving DeBERTa using ELECTRA-Style Pre-Training with
                  Gradient-Disentangled Embedding Sharing},
  booktitle    = {The Eleventh International Conference on Learning Representations,
                  {ICLR} 2023, Kigali, Rwanda, May 1-5, 2023},
  publisher    = {OpenReview.net},
  year         = {2023},
  url          = {https://openreview.net/forum?id=sE7-XhLxHA},
  bibsource    = {dblp computer science bibliography, https://dblp.org}
}

@article{nomic,
  author       = {Zach Nussbaum and
                  John X. Morris and
                  Brandon Duderstadt and
                  Andriy Mulyar},
  title        = {Nomic Embed: Training a Reproducible Long Context Text Embedder},
  journal      = {CoRR},
  volume       = {abs/2402.01613},
  year         = {2024},
  url          = {https://doi.org/10.48550/arXiv.2402.01613},
  doi          = {10.48550/ARXIV.2402.01613},
  eprinttype    = {arXiv},
  eprint       = {2402.01613},
  bibsource    = {dblp computer science bibliography, https://dblp.org}
}

@article{roberta,
  author       = {Yinhan Liu and
                  Myle Ott and
                  Naman Goyal and
                  Jingfei Du and
                  Mandar Joshi and
                  Danqi Chen and
                  Omer Levy and
                  Mike Lewis and
                  Luke Zettlemoyer and
                  Veselin Stoyanov},
  title        = {RoBERTa: {A} Robustly Optimized {BERT} Pretraining Approach},
  journal      = {CoRR},
  volume       = {abs/1907.11692},
  year         = {2019},
  url          = {http://arxiv.org/abs/1907.11692},
  eprinttype    = {arXiv},
  eprint       = {1907.11692},
  bibsource    = {dblp computer science bibliography, https://dblp.org}
}

@article{olmo,
  title={Olmo: Accelerating the science of language models},
  author={Groeneveld, Dirk and Beltagy, Iz and Walsh, Pete and Bhagia, Akshita and Kinney, Rodney and Tafjord, Oyvind and Jha, Ananya Harsh and Ivison, Hamish and Magnusson, Ian and Wang, Yizhong and others},
  journal={arXiv preprint arXiv:2402.00838},
  year={2024}
}

@inproceedings{MNLI,
  title={A Broad-Coverage Challenge Corpus for Sentence Understanding through Inference},
  author={Williams, Adina and Nangia, Nikita and Bowman, Samuel},
  booktitle={Proceedings of the 2018 Conference of the North American Chapter of the Association for Computational Linguistics: Human Language Technologies, Volume 1 (Long Papers)},
  pages={1112--1122},
  year={2018}
}

@inproceedings{pythia,
  title={Pythia: A suite for analyzing large language models across training and scaling},
  author={Biderman, Stella and Schoelkopf, Hailey and Anthony, Quentin Gregory and Bradley, Herbie and O’Brien, Kyle and Hallahan, Eric and Khan, Mohammad Aflah and Purohit, Shivanshu and Prashanth, USVSN Sai and Raff, Edward and others},
  booktitle={International Conference on Machine Learning},
  pages={2397--2430},
  year={2023},
  organization={PMLR}
}

@article{llama3,
  title={The llama 3 herd of models},
  author={Dubey, Abhimanyu and Jauhri, Abhinav and Pandey, Abhinav and Kadian, Abhishek and Al-Dahle, Ahmad and Letman, Aiesha and Mathur, Akhil and Schelten, Alan and Yang, Amy and Fan, Angela and others},
  journal={arXiv preprint arXiv:2407.21783},
  year={2024}
}

@inproceedings{mosaic,
  author       = {Jacob Portes and
                  Alexander Trott and
                  Sam Havens and
                  Daniel King and
                  Abhinav Venigalla and
                  Moin Nadeem and
                  Nikhil Sardana and
                  Daya Khudia and
                  Jonathan Frankle},
  editor       = {Alice Oh and
                  Tristan Naumann and
                  Amir Globerson and
                  Kate Saenko and
                  Moritz Hardt and
                  Sergey Levine},
  title        = {MosaicBERT: {A} Bidirectional Encoder Optimized for Fast Pretraining},
  booktitle    = {Advances in Neural Information Processing Systems 36: Annual Conference
                  on Neural Information Processing Systems 2023, NeurIPS 2023, New Orleans,
                  LA, USA, December 10 - 16, 2023},
  year         = {2023},
  url          = {http://papers.nips.cc/paper\_files/paper/2023/hash/095a6917768712b7ccc61acbeecad1d8-Abstract-Conference.html},
  bibsource    = {dblp computer science bibliography, https://dblp.org}
}

@misc{olmo2,
      title={2 OLMo 2 Furious}, 
      author={Team OLMo and Pete Walsh and Luca Soldaini and Dirk Groeneveld and Kyle Lo and Shane Arora and Akshita Bhagia and Yuling Gu and Shengyi Huang and Matt Jordan and Nathan Lambert and Dustin Schwenk and Oyvind Tafjord and Taira Anderson and David Atkinson and Faeze Brahman and Christopher Clark and Pradeep Dasigi and Nouha Dziri and Michal Guerquin and Hamish Ivison and Pang Wei Koh and Jiacheng Liu and Saumya Malik and William Merrill and Lester James V. Miranda and Jacob Morrison and Tyler Murray and Crystal Nam and Valentina Pyatkin and Aman Rangapur and Michael Schmitz and Sam Skjonsberg and David Wadden and Christopher Wilhelm and Michael Wilson and Luke Zettlemoyer and Ali Farhadi and Noah A. Smith and Hannaneh Hajishirzi},
      year={2025},
      eprint={2501.00656},
      archivePrefix={arXiv},
      primaryClass={cs.CL},
      url={https://arxiv.org/abs/2501.00656}, 
}

@article{msmarco,
  title={Ms marco: A human generated machine reading comprehension dataset},
  author={Bajaj, Payal and Campos, Daniel and Craswell, Nick and Deng, Li and Gao, Jianfeng and Liu, Xiaodong and Majumder, Rangan and McNamara, Andrew and Mitra, Bhaskar and Nguyen, Tri and others},
  journal={arXiv preprint arXiv:1611.09268},
  year={2016}
}

@inproceedings{gpt3,
  author       = {Tom B. Brown and
                  Benjamin Mann and
                  Nick Ryder and
                  Melanie Subbiah and
                  Jared Kaplan and
                  Prafulla Dhariwal and
                  Arvind Neelakantan and
                  Pranav Shyam and
                  Girish Sastry and
                  Amanda Askell and
                  Sandhini Agarwal and
                  Ariel Herbert{-}Voss and
                  Gretchen Krueger and
                  Tom Henighan and
                  Rewon Child and
                  Aditya Ramesh and
                  Daniel M. Ziegler and
                  Jeffrey Wu and
                  Clemens Winter and
                  Christopher Hesse and
                  Mark Chen and
                  Eric Sigler and
                  Mateusz Litwin and
                  Scott Gray and
                  Benjamin Chess and
                  Jack Clark and
                  Christopher Berner and
                  Sam McCandlish and
                  Alec Radford and
                  Ilya Sutskever and
                  Dario Amodei},
  editor       = {Hugo Larochelle and
                  Marc'Aurelio Ranzato and
                  Raia Hadsell and
                  Maria{-}Florina Balcan and
                  Hsuan{-}Tien Lin},
  title        = {Language Models are Few-Shot Learners},
  booktitle    = {Advances in Neural Information Processing Systems 33: Annual Conference
                  on Neural Information Processing Systems 2020, NeurIPS 2020, December
                  6-12, 2020, virtual},
  year         = {2020},
  url          = {https://proceedings.neurips.cc/paper/2020/hash/1457c0d6bfcb4967418bfb8ac142f64a-Abstract.html},
  bibsource    = {dblp computer science bibliography, https://dblp.org}
}

@article{sst2,
  author       = {Yinhan Liu and
                  Myle Ott and
                  Naman Goyal and
                  Jingfei Du and
                  Mandar Joshi and
                  Danqi Chen and
                  Omer Levy and
                  Mike Lewis and
                  Luke Zettlemoyer and
                  Veselin Stoyanov},
  title        = {RoBERTa: {A} Robustly Optimized {BERT} Pretraining Approach},
  journal      = {CoRR},
  volume       = {abs/1907.11692},
  year         = {2019},
  url          = {http://arxiv.org/abs/1907.11692},
  eprinttype    = {arXiv},
  eprint       = {1907.11692},
  bibsource    = {dblp computer science bibliography, https://dblp.org}
}

@misc{eval-harness,
  author       = {Gao, Leo and Tow, Jonathan and Abbasi, Baber and Biderman, Stella and Black, Sid and DiPofi, Anthony and Foster, Charles and Golding, Laurence and Hsu, Jeffrey and Le Noac'h, Alain and Li, Haonan and McDonell, Kyle and Muennighoff, Niklas and Ociepa, Chris and Phang, Jason and Reynolds, Laria and Schoelkopf, Hailey and Skowron, Aviya and Sutawika, Lintang and Tang, Eric and Thite, Anish and Wang, Ben and Wang, Kevin and Zou, Andy},
  title        = {The Language Model Evaluation Harness},
  month        = 07,
  year         = 2024,
  publisher    = {Zenodo},
  version      = {v0.4.3},
  doi          = {10.5281/zenodo.12608602},
  url          = {https://zenodo.org/records/12608602}
}

@article{marone2025mmbert,
  title={mmbert: A modern multilingual encoder with annealed language learning},
  author={Marone, Marc and Weller, Orion and Fleshman, William and Yang, Eugene and Lawrie, Dawn and Van Durme, Benjamin},
  journal={arXiv preprint arXiv:2509.06888},
  year={2025}
}

@misc{gisserotboukhlef2025pretrainencodersmaskedlanguage,
      title={Should We Still Pretrain Encoders with Masked Language Modeling?}, 
      author={Hippolyte Gisserot-Boukhlef and Nicolas Boizard and Manuel Faysse and Duarte M. Alves and Emmanuel Malherbe and André F. T. Martins and Céline Hudelot and Pierre Colombo},
      year={2025},
      eprint={2507.00994},
      archivePrefix={arXiv},
      primaryClass={cs.CL},
      url={https://arxiv.org/abs/2507.00994}, 
}

@inproceedings{ea01b9c0db064caca6986b925d75f2bb,
    title = "The winograd schema challenge",
    author = "Levesque, {Hector J.} and Ernest Davis and Leora Morgenstern",
    year = "2012",
    language = "English (US)",
    isbn = "9781577355601",
    series = "Proceedings of the International Conference on Knowledge Representation and Reasoning",
    publisher = "Institute of Electrical and Electronics Engineers Inc.",
    pages = "552--561",
    booktitle = "13th International Conference on the Principles of Knowledge Representation and Reasoning, KR 2012",
    note = "13th International Conference on the Principles of Knowledge Representation and Reasoning, KR 2012 ; Conference date: 10-06-2012 Through 14-06-2012",
}

@article{sakaguchi2021winogrande,
  title={Winogrande: An adversarial winograd schema challenge at scale},
  author={Sakaguchi, Keisuke and Bras, Ronan Le and Bhagavatula, Chandra and Choi, Yejin},
  journal={Communications of the ACM},
  volume={64},
  number={9},
  pages={99--106},
  year={2021},
  publisher={ACM New York, NY, USA}
}

@article{joshi2017triviaqa,
  title={Triviaqa: A large scale distantly supervised challenge dataset for reading comprehension},
  author={Joshi, Mandar and Choi, Eunsol and Weld, Daniel S and Zettlemoyer, Luke},
  journal={arXiv preprint arXiv:1705.03551},
  year={2017}
}

@article{sap2019socialiqa,
  title={Socialiqa: Commonsense reasoning about social interactions},
  author={Sap, Maarten and Rashkin, Hannah and Chen, Derek and LeBras, Ronan and Choi, Yejin},
  journal={arXiv preprint arXiv:1904.09728},
  year={2019}
}

@article{mihaylov2018can,
  title={Can a suit of armor conduct electricity? a new dataset for open book question answering},
  author={Mihaylov, Todor and Clark, Peter and Khot, Tushar and Sabharwal, Ashish},
  journal={arXiv preprint arXiv:1809.02789},
  year={2018}
}

@article{paperno2016lambada,
  title={The LAMBADA dataset: Word prediction requiring a broad discourse context},
  author={Paperno, Denis and Kruszewski, Germ{\'a}n and Lazaridou, Angeliki and Pham, Quan Ngoc and Bernardi, Raffaella and Pezzelle, Sandro and Baroni, Marco and Boleda, Gemma and Fern{\'a}ndez, Raquel},
  journal={arXiv preprint arXiv:1606.06031},
  year={2016}
}

@article{clark2018think,
  title={Think you have solved question answering? try arc, the ai2 reasoning challenge},
  author={Clark, Peter and Cowhey, Isaac and Etzioni, Oren and Khot, Tushar and Sabharwal, Ashish and Schoenick, Carissa and Tafjord, Oyvind},
  journal={arXiv preprint arXiv:1803.05457},
  year={2018}
}

@article{zellers2019hellaswag,
  title={Hellaswag: Can a machine really finish your sentence?},
  author={Zellers, Rowan and Holtzman, Ari and Bisk, Yonatan and Farhadi, Ali and Choi, Yejin},
  journal={arXiv preprint arXiv:1905.07830},
  year={2019}
}

@article{husain2019codesearchnet,
  title={Codesearchnet challenge: Evaluating the state of semantic code search},
  author={Husain, Hamel and Wu, Ho-Hsiang and Gazit, Tiferet and Allamanis, Miltiadis and Brockschmidt, Marc},
  journal={arXiv preprint arXiv:1909.09436},
  year={2019}
}

@misc{bge-m3,
      title={BGE M3-Embedding: Multi-Lingual, Multi-Functionality, Multi-Granularity Text Embeddings Through Self-Knowledge Distillation}, 
      author={Jianlv Chen and Shitao Xiao and Peitian Zhang and Kun Luo and Defu Lian and Zheng Liu},
      year={2024},
      eprint={2402.03216},
      archivePrefix={arXiv},
      primaryClass={cs.CL}
}

@article{wang2018glue,
  title={GLUE: A multi-task benchmark and analysis platform for natural language understanding},
  author={Wang, Alex and Singh, Amanpreet and Michael, Julian and Hill, Felix and Levy, Omer and Bowman, Samuel R},
  journal={arXiv preprint arXiv:1804.07461},
  year={2018}
}

@inproceedings{biderman2023pythia,
  title={Pythia: A suite for analyzing large language models across training and scaling},
  author={Biderman, Stella and Schoelkopf, Hailey and Anthony, Quentin Gregory and Bradley, Herbie and O’Brien, Kyle and Hallahan, Eric and Khan, Mohammad Aflah and Purohit, Shivanshu and Prashanth, USVSN Sai and Raff, Edward and others},
  booktitle={International Conference on Machine Learning},
  pages={2397--2430},
  year={2023},
  organization={PMLR}
}

@article{allal2025smollm2,
  title={SmolLM2: When Smol Goes Big--Data-Centric Training of a Small Language Model},
  author={Allal, Loubna Ben and Lozhkov, Anton and Bakouch, Elie and Bl{\'a}zquez, Gabriel Mart{\'\i}n and Penedo, Guilherme and Tunstall, Lewis and Marafioti, Andr{\'e}s and Kydl{\'\i}{\v{c}}ek, Hynek and Lajar{\'\i}n, Agust{\'\i}n Piqueres and Srivastav, Vaibhav and others},
  journal={arXiv preprint arXiv:2502.02737},
  year={2025}
}

@article{muennighoff2023scaling,
  title={Scaling data-constrained language models},
  author={Muennighoff, Niklas and Rush, Alexander and Barak, Boaz and Le Scao, Teven and Tazi, Nouamane and Piktus, Aleksandra and Pyysalo, Sampo and Wolf, Thomas and Raffel, Colin A},
  journal={Advances in Neural Information Processing Systems},
  volume={36},
  pages={50358--50376},
  year={2023}
}

@inproceedings{liu2024mobilellm,
  title={Mobilellm: Optimizing sub-billion parameter language models for on-device use cases},
  author={Liu, Zechun and Zhao, Changsheng and Iandola, Forrest and Lai, Chen and Tian, Yuandong and Fedorov, Igor and Xiong, Yunyang and Chang, Ernie and Shi, Yangyang and Krishnamoorthi, Raghuraman and others},
  booktitle={Forty-first International Conference on Machine Learning},
  year={2024}
}

@article{gao2024train,
  title={How to train long-context language models (effectively)},
  author={Gao, Tianyu and Wettig, Alexander and Yen, Howard and Chen, Danqi},
  journal={arXiv preprint arXiv:2410.02660},
  year={2024}
}

@article{su2024roformer,
  title={Roformer: Enhanced transformer with rotary position embedding},
  author={Su, Jianlin and Ahmed, Murtadha and Lu, Yu and Pan, Shengfeng and Bo, Wen and Liu, Yunfeng},
  journal={Neurocomputing},
  volume={568},
  pages={127063},
  year={2024},
  publisher={Elsevier}
}

@misc{lozhkov2024finewebedu,
    author       = { Lozhkov, Anton and Ben Allal, Loubna and von Werra, Leandro and Wolf, Thomas },  
    title        = { FineWeb-Edu: the Finest Collection of Educational Content }, 
    year         = 2024,  
    url          = { https://huggingface.co/datasets/HuggingFaceFW/fineweb-edu },  
    doi          = { 10.57967/hf/2497 },
    publisher    = { Hugging Face }
}

@article{penedo2024fineweb,
  title={The fineweb datasets: Decanting the web for the finest text data at scale},
  author={Penedo, Guilherme and Kydl{\'\i}{\v{c}}ek, Hynek and Lozhkov, Anton and Mitchell, Margaret and Raffel, Colin A and Von Werra, Leandro and Wolf, Thomas and others},
  journal={Advances in Neural Information Processing Systems},
  volume={37},
  pages={30811--30849},
  year={2024}
}

@article{nielsen2024encoder,
  title={Encoder vs decoder: Comparative analysis of encoder and decoder language models on multilingual nlu tasks},
  author={Nielsen, Dan Saattrup and Enevoldsen, Kenneth and Schneider-Kamp, Peter},
  journal={arXiv preprint arXiv:2406.13469},
  year={2024}
}

@article{soldaini2024dolma,
  title={Dolma: An open corpus of three trillion tokens for language model pretraining research},
  author={Soldaini, Luca and Kinney, Rodney and Bhagia, Akshita and Schwenk, Dustin and Atkinson, David and Authur, Russell and Bogin, Ben and Chandu, Khyathi and Dumas, Jennifer and Elazar, Yanai and others},
  journal={arXiv preprint arXiv:2402.00159},
  year={2024}
}

@article{li2024datacomp,
  title={Datacomp-lm: In search of the next generation of training sets for language models},
  author={Li, Jeffrey and Fang, Alex and Smyrnis, Georgios and Ivgi, Maor and Jordan, Matt and Gadre, Samir Yitzhak and Bansal, Hritik and Guha, Etash and Keh, Sedrick Scott and Arora, Kushal and others},
  journal={Advances in Neural Information Processing Systems},
  volume={37},
  pages={14200--14282},
  year={2024}
}

@article{rehana2023evaluation,
  title={Evaluation of GPT and BERT-based models on identifying proteinprotein interactions in biomedical text},
  author={Rehana, Hasin and {\c{C}}am, Nur Bengisu and Basmaci, Mert and Zheng, Jie and Jemiyo, Christianah and He, Yongqun and {\"O}zg{\"u}r, Arzucan and Hur, Junguk},
  journal={ArXiv},
  pages={arXiv--2303},
  year={2023}
}

@inproceedings{zheng2021adapting,
  title={Adapting GPT, GPT-2 and BERT language models for speech recognition},
  author={Zheng, Xianrui and Zhang, Chao and Woodland, Philip C},
  booktitle={2021 IEEE Automatic speech recognition and understanding workshop (ASRU)},
  pages={162--168},
  year={2021},
  organization={IEEE}
}

@inproceedings{qu2020text,
  title={A text generation and prediction system: pre-training on new corpora using BERT and GPT-2},
  author={Qu, Yuanbin and Liu, Peihan and Song, Wei and Liu, Lizhen and Cheng, Miaomiao},
  booktitle={2020 IEEE 10th international conference on electronics information and emergency communication (ICEIEC)},
  pages={323--326},
  year={2020},
  organization={IEEE}
}

@inproceedings{yang2023recent,
  title={Recent progress on text summarisation based on bert and gpt},
  author={Yang, Binxia and Luo, Xudong and Sun, Kaili and Luo, Michael Y},
  booktitle={International conference on knowledge science, engineering and management},
  pages={225--241},
  year={2023},
  organization={Springer}
}

@article{wang2020minilm,
  title={Minilm: Deep self-attention distillation for task-agnostic compression of pre-trained transformers},
  author={Wang, Wenhui and Wei, Furu and Dong, Li and Bao, Hangbo and Yang, Nan and Zhou, Ming},
  journal={Advances in neural information processing systems},
  volume={33},
  pages={5776--5788},
  year={2020}
}

@article{turc2019well,
  title={Well-read students learn better: On the importance of pre-training compact models},
  author={Turc, Iulia and Chang, Ming-Wei and Lee, Kenton and Toutanova, Kristina},
  journal={arXiv preprint arXiv:1908.08962},
  year={2019}
}

@article{sanh2019distilbert,
  title={DistilBERT, a distilled version of BERT: smaller, faster, cheaper and lighter},
  author={Sanh, Victor and Debut, Lysandre and Chaumond, Julien and Wolf, Thomas},
  journal={arXiv preprint arXiv:1910.01108},
  year={2019}
}

@article{jiao2019tinybert,
  title={Tinybert: Distilling bert for natural language understanding},
  author={Jiao, Xiaoqi and Yin, Yichun and Shang, Lifeng and Jiang, Xin and Chen, Xiao and Li, Linlin and Wang, Fang and Liu, Qun},
  journal={arXiv preprint arXiv:1909.10351},
  year={2019}
}

@article{grattafiori2024llama,
  title={The llama 3 herd of models},
  author={Grattafiori, Aaron and Dubey, Abhimanyu and Jauhri, Abhinav and Pandey, Abhinav and Kadian, Abhishek and Al-Dahle, Ahmad and Letman, Aiesha and Mathur, Akhil and Schelten, Alan and Vaughan, Alex and others},
  journal={arXiv preprint arXiv:2407.21783},
  year={2024}
}

@article{yang2025qwen3,
  title={Qwen3 technical report},
  author={Yang, An and Li, Anfeng and Yang, Baosong and Zhang, Beichen and Hui, Binyuan and Zheng, Bo and Yu, Bowen and Gao, Chang and Huang, Chengen and Lv, Chenxu and others},
  journal={arXiv preprint arXiv:2505.09388},
  year={2025}
}

@article{achiam2023gpt,
  title={Gpt-4 technical report},
  author={Achiam, Josh and Adler, Steven and Agarwal, Sandhini and Ahmad, Lama and Akkaya, Ilge and Aleman, Florencia Leoni and Almeida, Diogo and Altenschmidt, Janko and Altman, Sam and Anadkat, Shyamal and others},
  journal={arXiv preprint arXiv:2303.08774},
  year={2023}
}

@article{team2023gemini,
  title={Gemini: a family of highly capable multimodal models},
  author={Team, Gemini and Anil, Rohan and Borgeaud, Sebastian and Alayrac, Jean-Baptiste and Yu, Jiahui and Soricut, Radu and Schalkwyk, Johan and Dai, Andrew M and Hauth, Anja and Millican, Katie and others},
  journal={arXiv preprint arXiv:2312.11805},
  year={2023}
}

@article{radford2019language,
  title={Language models are unsupervised multitask learners},
  author={Radford, Alec and Wu, Jeffrey and Child, Rewon and Luan, David and Amodei, Dario and Sutskever, Ilya and others},
  journal={OpenAI blog},
  volume={1},
  number={8},
  pages={9},
  year={2019}
}

@article{zhang2024min,
  title={Min-k\%++: Improved baseline for detecting pre-training data from large language models},
  author={Zhang, Jingyang and Sun, Jingwei and Yeats, Eric and Ouyang, Yang and Kuo, Martin and Zhang, Jianyi and Yang, Hao Frank and Li, Hai},
  journal={arXiv preprint arXiv:2404.02936},
  year={2024}
}

@article{cheng2024dated,
  title={Dated data: Tracing knowledge cutoffs in large language models},
  author={Cheng, Jeffrey and Marone, Marc and Weller, Orion and Lawrie, Dawn and Khashabi, Daniel and Van Durme, Benjamin},
  journal={arXiv preprint arXiv:2403.12958},
  year={2024}
}

@article{oh2024frequency,
  title={Frequency Explains the Inverse Correlation of Large Language Models' Size, Training Data Amount, and Surprisal's Fit to Reading Times},
  author={Oh, Byung-Doh and Yue, Shisen and Schuler, William},
  journal={arXiv preprint arXiv:2402.02255},
  year={2024}
}

@inproceedings{longpre2024pretrainer,
  title={A pretrainer’s guide to training data: Measuring the effects of data age, domain coverage, quality, \& toxicity},
  author={Longpre, Shayne and Yauney, Gregory and Reif, Emily and Lee, Katherine and Roberts, Adam and Zoph, Barret and Zhou, Denny and Wei, Jason and Robinson, Kevin and Mimno, David and others},
  booktitle={Proceedings of the 2024 Conference of the North American Chapter of the Association for Computational Linguistics: Human Language Technologies (Volume 1: Long Papers)},
  pages={3245--3276},
  year={2024}
}

@article{behnamghader2024llm2vec,
  title={Llm2vec: Large language models are secretly powerful text encoders},
  author={BehnamGhader, Parishad and Adlakha, Vaibhav and Mosbach, Marius and Bahdanau, Dzmitry and Chapados, Nicolas and Reddy, Siva},
  journal={arXiv preprint arXiv:2404.05961},
  year={2024}
}

@article{lee2025clinical,
  title={Clinical modernbert: An efficient and long context encoder for biomedical text},
  author={Lee, Simon A and Wu, Anthony and Chiang, Jeffrey N},
  journal={arXiv preprint arXiv:2504.03964},
  year={2025}
}

@article{qwen3embedding,
  title={Qwen3 Embedding: Advancing Text Embedding and Reranking Through Foundation Models},
  author={Zhang, Yanzhao and Li, Mingxin and Long, Dingkun and Zhang, Xin and Lin, Huan and Yang, Baosong and Xie, Pengjun and Yang, An and Liu, Dayiheng and Lin, Junyang and Huang, Fei and Zhou, Jingren},
  journal={arXiv preprint arXiv:2506.05176},
  year={2025}
}

@article{harrag2021bert,
  title={Bert transformer model for detecting Arabic GPT2 auto-generated tweets},
  author={Harrag, Fouzi and Debbah, Maria and Darwish, Kareem and Abdelali, Ahmed},
  journal={arXiv preprint arXiv:2101.09345},
  year={2021}
}

@article{ethayarajh2019contextual,
  title={How contextual are contextualized word representations? Comparing the geometry of BERT, ELMo, and GPT-2 embeddings},
  author={Ethayarajh, Kawin},
  journal={arXiv preprint arXiv:1909.00512},
  year={2019}
}

@article{enevoldsen2025mmteb,
  title={Mmteb: Massive multilingual text embedding benchmark},
  author={Enevoldsen, Kenneth and Chung, Isaac and Kerboua, Imene and Kardos, M{\'a}rton and Mathur, Ashwin and Stap, David and Gala, Jay and Siblini, Wissam and Krzemi{\'n}ski, Dominik and Winata, Genta Indra and others},
  journal={arXiv preprint arXiv:2502.13595},
  year={2025}
}

@inproceedings{weller2025mfollowir,
  title={mFollowIR: A Multilingual Benchmark for Instruction Following in Retrieval},
  author={Weller, Orion and Chang, Benjamin and Yang, Eugene and Yarmohammadi, Mahsa and Barham, Samuel and MacAvaney, Sean and Cohan, Arman and Soldaini, Luca and Van Durme, Benjamin and Lawrie, Dawn},
  booktitle={European Conference on Information Retrieval},
  pages={295--310},
  year={2025}
}

@article{weller2024followir,
  title={Followir: Evaluating and teaching information retrieval models to follow instructions},
  author={Weller, Orion and Chang, Benjamin and MacAvaney, Sean and Lo, Kyle and Cohan, Arman and Van Durme, Benjamin and Lawrie, Dawn and Soldaini, Luca},
  journal={arXiv preprint arXiv:2403.15246},
  year={2024}
}

@article{shao2025reasonir,
  title={ReasonIR: Training Retrievers for Reasoning Tasks},
  author={Shao, Rulin and Qiao, Rui and Kishore, Varsha and Muennighoff, Niklas and Lin, Xi Victoria and Rus, Daniela and Low, Bryan Kian Hsiang and Min, Sewon and Yih, Wen-tau and Koh, Pang Wei and others},
  journal={arXiv preprint arXiv:2504.20595},
  year={2025}
}

@article{hoffmann2022training,
  title={Training compute-optimal large language models},
  author={Hoffmann, Jordan and Borgeaud, Sebastian and Mensch, Arthur and Buchatskaya, Elena and Cai, Trevor and Rutherford, Eliza and Casas, Diego de Las and Hendricks, Lisa Anne and Welbl, Johannes and Clark, Aidan and others},
  journal={arXiv preprint arXiv:2203.15556},
  year={2022}
}

@inproceedings{Du2021GLMGL,
  title={GLM: General Language Model Pretraining with Autoregressive Blank Infilling},
  author={Zhengxiao Du and Yujie Qian and Xiao Liu and Ming Ding and Jiezhong Qiu and Zhilin Yang and Jie Tang},
  booktitle={Annual Meeting of the Association for Computational Linguistics},
  year={2021},
  url={https://api.semanticscholar.org/CorpusID:247519241}
}

@article{chowdhery2023palm,
  title={Palm: Scaling language modeling with pathways},
  author={Chowdhery, Aakanksha and Narang, Sharan and Devlin, Jacob and Bosma, Maarten and Mishra, Gaurav and Roberts, Adam and Barham, Paul and Chung, Hyung Won and Sutton, Charles and Gehrmann, Sebastian and others},
  journal={Journal of Machine Learning Research},
  volume={24},
  number={240},
  pages={1--113},
  year={2023}
}

@article{artetxe2022role,
  title={On the role of bidirectionality in language model pre-training},
  author={Artetxe, Mikel and Du, Jingfei and Goyal, Naman and Zettlemoyer, Luke and Stoyanov, Ves},
  journal={arXiv preprint arXiv:2205.11726},
  year={2022}
}

@article{kopiczko2024bitune,
  title={Bitune: Bidirectional Instruction-Tuning},
  author={Kopiczko, Dawid J and Blankevoort, Tijmen and Asano, Yuki M},
  journal={arXiv preprint arXiv:2405.14862},
  year={2024}
}

@article{rudinger2018gender,
  title={Gender bias in coreference resolution},
  author={Rudinger, Rachel and Naradowsky, Jason and Leonard, Brian and Van Durme, Benjamin},
  journal={arXiv preprint arXiv:1804.09301},
  year={2018}
}

@article{warner2024smarter,
  title={Smarter, better, faster, longer: A modern bidirectional encoder for fast, memory efficient, and long context finetuning and inference},
  author={Warner, Benjamin and Chaffin, Antoine and Clavi{\'e}, Benjamin and Weller, Orion and Hallstr{\"o}m, Oskar and Taghadouini, Said and Gallagher, Alexis and Biswas, Raja and Ladhak, Faisal and Aarsen, Tom and others},
  journal={arXiv preprint arXiv:2412.13663},
  year={2024}
}

@article{samuel2024berts,
  title={BERTs are generative in-context learners},
  author={Samuel, David},
  journal={Advances in Neural Information Processing Systems},
  volume={37},
  pages={2558--2589},
  year={2024}
}

@inproceedings{charpentier-samuel-2024-bert,
    title = "{GPT} or {BERT}: why not both?",
    author = "Charpentier, Lucas Georges Gabriel  and
      Samuel, David",
    editor = "Hu, Michael Y.  and
      Mueller, Aaron  and
      Ross, Candace  and
      Williams, Adina  and
      Linzen, Tal  and
      Zhuang, Chengxu  and
      Choshen, Leshem  and
      Cotterell, Ryan  and
      Warstadt, Alex  and
      Wilcox, Ethan Gotlieb",
    booktitle = "The 2nd BabyLM Challenge at the 28th Conference on Computational Natural Language Learning",
    month = nov,
    year = "2024",
    address = "Miami, FL, USA",
    publisher = "Association for Computational Linguistics",
    url = "https://aclanthology.org/2024.conll-babylm.24/",
    pages = "262--283"
}
\bibliographystyle{iclr2026_conference}

\appendix

\section{All Encoder Results}
\label{app:all_encoder}
We show results for the full  MTEB v2 eng and GLUE results in Table~\ref{tab:mteb_results} and Table~\ref{tab:glue_results}  respectively.

\section{Decoder Evaluation Frameworks}
\label{app:eval}
For generative tasks, we use the Eleuther AI harness with commit \texttt{867413f8677f00f6a817262727cbb041bf36192a}. We also use a forked version of the Eleuther AI harness for evaluating encoders in a generative fashion (see Github for details). Following previous work \citep{allal2025smollm2} the ARC score is the average of the easy and challenge sets.

For encoders evaluated on generative tasks we use three mask tokens followed by the EOS token. At each step, we predict the first MASK token and iteratively generate. However, this approach still could be improved, in particular around the EOS token. Encoder models are non-calibrated for when this should appear, so we had to make small changes to this setup for two tasks: for TriviaQA we change the EOS token to a newline character (as the harness stops on newlines for TriviaQA also) and for the Lambada OpenAI task we do not score the EOS token. All other tasks proceed with the three masks + EOS token as proposed by \citet{samuel2024berts}.

\section{Architecture Details}
\label{app:arch}
Architecture and training details for all models are found in 
Table~\ref{tab:common}. These are generally the same as ModernBERT except for the same value of local and global RoPE and a slightly shorter context length (7999).

\section{Model Sizes}
\label{app:sizes}
Model sizes (both embedding and non-embedding) are found in Table~\ref{tab:model_params}. We group models by total parameters, although we note that some have more vocab parameters vs non-vocab parameters, e.g. MiniLM L12 has almost 2x the number of non-embedding parameters compared to \modelname-32m (21M vs 12M).

\section{Compute Configuration}
\label{app:compute}
We train the models on a comparatively small compute cluster. We train each model on a 4xH100 node using NVLink. The pre-training phase (the longest) takes approximately 6 days for the 17M model. The longest was for the 1B model, which we trained for approximately 40 days. Unfortunately, we did not have enough compute availability to train the 1B to the full 2T tokens. Thus the 1B models are scaled to 1/3 of the data (e.g. 667B instead of 2T tokens). However, this is still more than chinchilla optimal \citep{hoffmann2022training} and it still outperformed other 1B models trained longer.

\section{Encoder Evaluation Sweep Parameters}
\label{app:encoder_sweep}
Below we detail the sweep hyperparameters for the retrieval and classification tasks that require fine-tuning.

\paragraph{GLUE}
We re-use ModernBERT's evaluation setup but slightly increase the learning rate sweep in order to better fit the smaller parameter models (which typically use higher LRs). As the best LRs chosen by BERT and ModernBERT were lower than these, it does not affect their scores. We sweep for learning rates over $\{1\text{e-}5, 3\text{e-}5, 5\text{e-}5, 8\text{e-}5, 1\text{e-}4\}$, weight decay values over $\{1\text{e-}6, 5\text{e-}6, 8\text{e-}6, 1\text{e-}5\}$, and batch sizes over $\{16, 32\}$. We also sweep over epochs $\{1, 2, 3, 4\}$ if task $\in \{\text{mnli}, \text{sst2}, \text{rte}\}$, otherwise $\{2, 5, 10, 12\}$. This was a total of 160 sweeps per model, of which we select the best score per task per model to report. We start from the best MNLI checkpoint for fine-tuning on RTE, STS-B, and MRPC, following ModernBERT.

\paragraph{Retrieval} We sweep four LRs ($\{1\text{e-}4, 3\text{e-}4, 5\text{e-}4, 7\text{e-}4\}$) on MS MARCO dev and choose the best performing one to evaluate on the other retrieval datasets. We use a new retrieval training script due to being unable to exactly reproduce ModernBERT's precise scores. While doing so, we also improve the training process for all models due to the use of more negatives in training, achieving higher scores than that in the ModernBERT paper (which was not trying to optimize scores, but shows generally that our training script is effective). We trained with an effective batch size of 1024 with 4 accumulation steps. For DeBERTa-v2 it diverged for all learning rates we tried. Thus, to get it to converge, we changed the warmup to 20\% from 5\% and lowered the learning rate to 1e-5.

As \modelname\ models have been trained with instructions during pre-training, it is likely they are also more capable for instruction-based retrieval \citep{shao2025reasonir,weller2024followir,weller2025mfollowir}, however, we leave that for future work.

\section{LLM Usage}
LLMs were not used for paper writing, but were used for coding assistance and title brainstorming. All code was human verified.

\begin{table*}[htb]
\resizebox{\textwidth}{!}{
\centering
\begin{tabular}{lcc|ccccccc}
\toprule
\textbf{Model Name} & \textbf{Mean (Task)} & \textbf{Mean (Type)} & \textbf{Class.} & \textbf{Clus.} & \textbf{Pair. Class.} & \textbf{Rerank.} & \textbf{Retrieval} & \textbf{STS} & \textbf{Summ.} \\
\midrule
\multicolumn{10}{c}{\textbf{XXS Models (11-25M parameters)}} \\
\midrule
TinyBERT & 52.1 & 49.7 & 64.6 & 37.4 & 78.1 & 41.9 & 33.3 & 71.9 & 20.6 \\
BERT-mini & 51.8 & 49.2 & 61.7 & 39.0 & 77.8 & 41.5 & 34.7 & 70.8 & 19.3 \\
\modelname-Enc-17m & 52.3 & 48.9 & 63.3 & 39.1 & 74.8 & 42.0 & 35.6 & 71.6 & 16.0 \\
\midrule
\multicolumn{10}{c}{\textbf{XS Models (28-33M parameters)}} \\
\midrule
BERT-small & 54.0 & 51.1 & 64.7 & 39.6 & 79.4 & 41.7 & 38.1 & 73.0 & 21.6 \\
MiniLM L12 & 53.9 & 51.3 & 64.9 & 37.8 & 79.8 & 43.5 & 38.4 & 73.2 & 21.4 \\
\modelname-Enc-32m & 54.2 & 50.9 & 64.2 & 39.6 & 77.1 & 42.6 & 39.7 & 73.2 & 19.9 \\
\midrule
\multicolumn{10}{c}{\textbf{S Models (68-82M parameters)}} \\
\midrule
DistilBERT & 55.5 & 52.7 & 66.5 & 39.8 & 80.7 & 42.8 & 40.8 & 74.1 & 24.0 \\
DistilRoBa & 55.2 & 51.8 & 67.3 & 39.3 & 78.9 & 43.5 & 40.0 & 74.2 & 19.4 \\
\modelname-Enc-68m & 56.1 & 52.6 & 66.6 & 40.1 & 79.3 & 43.3 & 43.1 & 74.2 & 21.7 \\
\midrule
\multicolumn{10}{c}{\textbf{Base Models (86-150M parameters)}} \\
\midrule
BERT-base & 56.0 & 52.9 & 67.2 & 40.4 & 80.5 & 43.1 & 41.2 & 74.8 & 23.1 \\
ModernBERT-base & 57.1 & 54.0 & 67.5 & 41.3 & 80.4 & 44.7 & 43.9 & 75.3 & 25.2 \\
\modelname-Enc-150m & 57.7 & 54.0 & 68.6 & 41.5 & 80.2 & 44.7 & 45.7 & 74.9 & 22.6 \\
\midrule
\multicolumn{10}{c}{\textbf{Large Models (305-395M parameters)}} \\
\midrule
BERT-large & 57.2 & 53.8 & 68.3 & 41.5 & 81.1 & 44.3 & 42.9 & 76.1 & 22.5 \\
ModernBERT-large & 58.6 & 55.0 & 69.1 & 41.5 & 82.2 & 45.5 & 47.0 & 76.5 & 23.5 \\
\modelname-Enc-400m & 59.4 & 55.5 & 69.9 & 41.8 & 82.6 & 45.6 & 48.4 & 77.2 & 22.6 \\
\midrule
\multicolumn{10}{c}{\textbf{XL Models (750-1565M parameters)}} \\
\midrule
DeBa-v1-xl & 59.5 & 56.4 & 70.8 & 42.5 & 82.7 & 45.7 & 47.2 & 77.1 & 28.6 \\
DeBa-v2-xxl* & 60.5 & 57.4 & 71.7 & 44.4 & 82.4 & 46.5 & 47.7 & 78.3 & 30.9 \\
\modelname-Enc-1b & 60.4 & 56.0 & 72.2 & 41.9 & 83.3 & 46.4 & 50.1 & 77.7 & 20.4 \\
\bottomrule
\end{tabular}
}
\caption{MTEB v2 English results. Class. = Classification, Clus. = Clustering, Pair. Class. = Pair Classification, Rerank. = Reranking, STS = Semantic Textual Similarity, Summ. = Summarization. DeBERTa v2 XXL did not converge with the standard learning rate sweeps, so we used a lower learning rate in order to help it to converge.}
\label{tab:mteb_results}
\end{table*}

\begin{table*}[htb]
\resizebox{\textwidth}{!}{
\centering
\begin{tabular}{l|cccccccc|c}
\toprule
 & \multicolumn{2}{c}{Single Sentence} & \multicolumn{3}{c}{Paraphrase and Similarity} & \multicolumn{3}{c|}{Natural Language Inference} \\
\cmidrule(lr){2-3} \cmidrule(lr){4-6} \cmidrule(lr){7-9}
\textbf{Model Name} & CoLA & SST-2 & MRPC & STS-B & QQP & MNLI & QNLI & RTE & \textbf{Avg} \\
\midrule
\multicolumn{10}{c}{\textbf{XXS Models (11-25M parameters)}} \\
\midrule
BERT-mini & 33.9 & 88.3 & 83.8 & 86.4 & 89.3 & 77.2 & 85.4 & 67.1 & 76.4 \\
TinyBERT & 22.4 & 91.2 & 87.5 & 87.8 & 89.4 & 80.9 & 88.4 & 68.2 & 77.0 \\
\modelname-Enc-17m & 43.9 & 91.2 & 86.0 & 87.2 & 89.8 & 79.5 & 87.3 & 69.0 & 79.2 \\
\midrule
\multicolumn{10}{c}{\textbf{XS Models (28-33M parameters)}} \\
\midrule
BERT-small & 44.8 & 90.1 & 83.1 & 87.6 & 90.1 & 79.2 & 88.2 & 69.3 & 79.0 \\
MiniLM L12 & 59.1 & 93.3 & 91.2 & 89.2 & 91.5 & 85.6 & 91.9 & 74.7 & 84.6 \\
\modelname-Enc-32m & 57.4 & 92.0 & 89.7 & 89.5 & 91.0 & 83.4 & 90.7 & 74.7 & 83.5 \\
\midrule
\multicolumn{10}{c}{\textbf{S Models (68-82M parameters)}} \\
\midrule
DistilBERT & 56.9 & 92.2 & 86.8 & 87.4 & 90.8 & 82.7 & 89.5 & 66.1 & 81.5 \\
DistilRoBERTa & 61.9 & 93.1 & 89.0 & 88.9 & 91.5 & 84.7 & 91.7 & 69.7 & 83.8 \\
\modelname-Enc-68m & 64.8 & 94.4 & 92.2 & 91.1 & 91.9 & 87.0 & 92.9 & 83.8 & 87.2 \\
\midrule
\multicolumn{10}{c}{\textbf{Base Models (86-150M parameters)}} \\
\midrule
BERT-base & 59.0 & 93.1 & 89.5 & 89.4 & 91.4 & 85.4 & 91.6 & 78.2 & 84.7 \\
ModernBERT-base & 65.1 & 96.0 & 92.2 & 91.8 & 92.1 & 89.1 & 93.9 & 87.4 & 88.4 \\
\modelname-Enc-150m & 66.9 & 95.8 & 92.6 & 92.2 & 92.4 & 89.2 & 94.0 & 87.7 & 88.9 \\
\midrule
\multicolumn{10}{c}{\textbf{Large Models (305-395M parameters)}} \\
\midrule
BERT-large & 56.2 & 93.3 & 87.8 & 90.6 & 90.9 & 86.3 & 92.8 & 83.8 & 85.2 \\
ModernBERT-large & 71.4 & 97.1 & 91.7 & 92.8 & 92.7 & 90.8 & 95.2 & 92.1 & 90.4 \\ 
\modelname-Enc-400m & 71.3 & 96.7 & 93.6 & 92.7 & 93.0 & 91.3 & 95.2 & 92.8 & 90.8 \\
\midrule
\multicolumn{10}{c}{\textbf{XL Models (750-1565M parameters)}} \\
\midrule
DeBERTa-v2-XL & 75.3 & 97.1 & 91.7 & 92.5 & 92.6 & 91.7 & 95.9 & 89.2 & 90.7 \\
DeBERTa-v2-XXL & 71.6 & - & - & - & - & 91.2 & 96.0 & - & - \\
\modelname-Enc-1b & 74.4 & 97.1 & 94.4 & 93.2 & 93.0 & 91.8 & 96.0 & 93.1 & 91.6 \\
\bottomrule
\end{tabular}
}
\caption{GLUE benchmark results across model sizes and architectures. DeBERTa v2 XXL was run for 300+ GPU hours before ruling it out due to it's large size (> 1.5B), results are incomplete.}
\label{tab:glue_results}
\end{table*}

\begin{table*}[htb]
\resizebox{\textwidth}{!}{
\centering
\begin{tabular}{l|cccccccccc|c}
\toprule
\textbf{Model Name} & \textbf{ARC} & \textbf{HS} & \textbf{LMB} & \textbf{OBQA} & \textbf{PIQA} & \textbf{SIQA} & \textbf{SciQ} & \textbf{TQA} & \textbf{WG} & \textbf{WSC} & \textbf{Avg} \\
\midrule
\multicolumn{12}{c}{\textbf{XXS Models (17M parameters)}} \\
\midrule
\modelname-Enc-from-Dec-17m & 27.7 & 27.2 & 23.4 & 31.4 & 56.0 & 34.6 & 45.9 & 0.5 & 51.2 & 52.7 & 35.1 \\
\modelname-Enc-17m & 28.3 & 26.4 & 24.1 & 34.0 & 54.2 & 34.4 & 44.0 & 0.1 & 52.6 & 52.7 & 35.1 \\
\modelname-Dec-From-Enc-17m &  22.7 & 26.8 & 21.9 & 24.6 & 56.1 & 70.9 & 35.7 & 0.9 & 53.4 & 53.8 & 36.7 \\
\modelname-Dec-17m & 21.3 & 27.1 & 23.0 & 27.2 & 57.7 & 71.1 & 35.4 & 2.6 & 50.9 & 48.0 & 36.4 \\
\midrule
\multicolumn{12}{c}{\textbf{XS Models (32M parameters)}} \\
\midrule
\modelname-Enc-from-Dec-32m & 28.2 & 27.9 & 29.5 & 33.8 & 55.6 & 34.7 & 45.6 & 0.1 & 53.2 & 50.9 & 36.0 \\
\modelname-Enc-32m & 28.7 & 28.0 & 33.6 & 34.8 & 56.7 & 34.4 & 41.4 & 0.2 & 51.4 & 56.4 & 36.6 \\
\modelname-Dec-From-Enc-32M & 20.5 & 28.3 & 27.7 & 27.0 & 58.1 & 77.2 & 36.0 & 3.0 & 50.2 & 52.7 & 38.1 \\
\modelname-Dec-32m  & 23.5 & 28.5 & 28.5 & 28.2 & 57.7 & 77.5 & 36.4 & 3.8 & 53.1 & 50.2 & 38.7 \\

\midrule
\multicolumn{12}{c}{\textbf{Small Models (68M parameters)}} \\
\midrule
\modelname-Enc-from-Dec-68m & 30.0 & 30.4 & 31.8 & 33.6 & 57.1 & 36.1 & 55.3 & 1.9 & 51.1 & 52.7 & 38.0 \\
\modelname-Enc-68m & 29.5 & 31.6 & 36.1 & 35.4 & 58.4 & 35.6 & 49.6 & 1.1 & 51.3 & 62.6 & 39.1 \\
\modelname-Dec-from-Enc-68m & 24.8 & 31.9 & 35.8 & 29.4 & 60.7 & 84.6 & 38.3 & 5.8 & 53.1 & 56.0 & 42.1 \\
\modelname-Dec-68m  & 25.3 & 33.4 & 35.2 & 29.4 & 61.8 & 83.2 & 38.8 & 5.6 & 50.1 & 55.3 & 41.8 \\
\midrule
\multicolumn{12}{c}{\textbf{Base Models (150M parameters)}} \\
\midrule
\modelname-Enc-from-Dec-150m & 33.5 & 36.3 & 39.2 & 34.4 & 63.9 & 39.7 & 74.3 & 4.7 & 51.5 & 59.3 & 43.7 \\
\modelname-Enc-150m & 32.5 & 36.5 & 41.6 & 37.4 & 63.0 & 38.5 & 59.8 & 1.6 & 54.9 & 63.0 & 42.9 \\
\modelname-Dec-from-Enc-150m & 25.0 & 36.0 & 39.4 & 30.0 & 62.9 & 84.7 & 40.4 & 7.4 & 52.9 & 57.5 & 43.6 \\
\modelname-Dec-150m  & 28.6 & 40.3 & 43.2 & 29.2 & 66.6 & 89.6 & 40.1 & 11.2 & 53.7 & 59.0 & 46.2 \\
\midrule
\multicolumn{12}{c}{\textbf{Large Models (400M parameters)}} \\
\midrule
\modelname-Enc-from-Dec-400m & 39.7 & 47.7 & 44.9 & 38.6 & 66.3 & 43.4 & 70.4 & 7.7 & 56.4 & 68.9 & 48.4 \\
\modelname-Enc-400m & 35.6 & 46.8 & 50.5 & 38.0 & 64.7 & 43.9 & 65.6 & 6.4 & 59.7 & 70.7 & 48.2 \\
\modelname--Dec-from-Enc-400m & 29.9 & 45.8 & 46.4 & 33.6 & 66.9 & 92.1 & 45.3 & 13.3 & 53.9 & 63.7 & 49.1 \\
\modelname-Dec-400m & 33.6 & 54.3 & 52.3 & 34.4 & 71.0 & 91.8 & 45.5 & 18.3 & 57.6 & 71.8 & 53.1 \\
\midrule
\multicolumn{12}{c}{\textbf{XL Models (1B parameters)}} \\
\midrule
\modelname-Enc-from-Dec-1B & 42.4 & 53.0 & 49.3 & 39.2 & 70.0 & 46.3 & 74.9 & 14.9 & 62.3 & 73.3 & 52.5 \\
\modelname-Enc-1B & 37.3 & 52.3 & 54.0 & 38.4 & 67.6 & 46.3 & 64.5 & 7.6 & 63.2 & 75.8 & 50.7 \\
\modelname-Dec-from-Enc-1B & 32.5 & 52.5 & 49.1 & 35.8 & 69.9 & 93.1 & 48.5 & 13.1 & 58.6 & 69.2 & 52.2 \\
\modelname-Dec-1B  & 39.7 & 62.9 & 58.4 & 41.6 & 74.4 & 93.8 & 48.2 & 29.3 & 62.7 & 79.1 & 59.0 \\
\bottomrule
\end{tabular}
}
\caption{Performance comparison of all models evaluated on generative tasks. Enc-from-Dec are trained with MTNP from decoders, while Dec-from-Enc are encoders trained with CLM.}
\label{tab:all_generative_results}
\end{table*}

\begin{table*}[htb]
\resizebox{\textwidth}{!}{
\centering
\begin{tabular}{l|ccc}
\toprule
\textbf{Model Name} & \textbf{Retrieval (nDCG@10)} & \textbf{MNLI (Accuracy)} & \textbf{Generative Avg} \\
\midrule
\multicolumn{4}{c}{\textbf{XXS Models (17M parameters)}} \\
\midrule
\modelname-Enc-17m & 30.93 & 79.5 & 35.1 \\
\modelname-Dec-17m & 29.11 & 77.6 & 36.4 \\
\modelname-Enc-from-Dec-17m & 31.01 & 77.7 & 35.1 \\
\modelname-Dec-from-Enc-17m & 28.52 & 78.8 & 36.7 \\
\midrule
\multicolumn{4}{c}{\textbf{XS Models (32M parameters)}} \\
\midrule
\modelname-Enc-32m & 35.13 & 83.4 & 36.6 \\
\modelname-Dec-32m & 32.93 & 80.4 & 38.7 \\
\modelname-Enc-from-Dec-32m & 34.66 & 80.9 & 36.0 \\
\modelname-Dec-from-Enc-32m & 32.32 & 82.6 & 38.1 \\
\midrule
\multicolumn{4}{c}{\textbf{Small Models (66-70M parameters)}} \\
\midrule
\modelname-Enc-68m & 38.17 & 87.0 & 39.1 \\
\modelname-Dec-68m & 36.12 & 83.9 & 41.8 \\
\modelname-Enc-from-Dec-68m & 37.87 & 83.9 & 38.0 \\
\modelname-Dec-from-Enc-68m & 36.31 & 85.8 & 42.1 \\
\midrule
\multicolumn{4}{c}{\textbf{Medium Models (150M parameters)}} \\
\midrule
\modelname-Enc-150m & 39.97 & 89.2 & 42.9 \\
\modelname-Dec-150m & 37.71 & 85.6 & 46.2 \\
\modelname-Enc-from-Dec-150m & 39.49 & 85.8 & 43.7 \\
\modelname-Dec-from-Enc-150m & 37.55 & 86.8 & 43.6 \\
\midrule
\multicolumn{4}{c}{\textbf{Large Models (400M parameters)}} \\
\midrule
\modelname-Enc-400m & 42.24 & 91.3 & 48.2 \\
\modelname-Dec-400m & 39.93 & 88.2 & 53.1 \\
\modelname-Enc-from-Dec-400m & 41.44 & 87.6 & 48.4 \\
\modelname-Dec-from-Enc-400m & 39.69 & 89.4 & 49.1 \\
\midrule
\multicolumn{4}{c}{\textbf{XL Models (1B parameters)}} \\
\midrule
\modelname-Enc-1b & 43.35 & 91.8 & 50.7 \\
\modelname-Dec-1b & 41.70 & 89.9 & 59.0 \\
\modelname-Enc-from-Dec-1b & 43.24 & 89.0 & 52.5 \\
\modelname-Dec-from-Enc-1b & 40.77 & 90.5 & 52.2 \\
\bottomrule
\end{tabular}
}
\caption{Table version of Figure~\ref{fig:enc_v_dec}. The generative eval breakdowns can be found in Table~\ref{tab:all_generative_results}.}
\label{tab:enc_vs_dec}
\end{table*}

\begin{table*}[htbp]
\tiny
\centering
\label{tab:winogender_acc}
\begin{tabular}{l|ccccccc}
\toprule
 & \multicolumn{4}{c}{\textbf{WinoGender All}} & \multicolumn{3}{c}{\textbf{WinoGender Gotcha}} \\
\cmidrule(lr){2-5} \cmidrule(lr){6-8}
\textbf{Model Name} & \textbf{Overall} & \textbf{Female} & \textbf{Male} & \textbf{Neutral} & \textbf{Overall} & \textbf{Female} & \textbf{Male} \\
\midrule
\multicolumn{8}{c}{\textbf{XXS Models (17M parameters)}} \\
\midrule
\modelname-Enc-from-Dec-17m & 50.8 $\pm$ 1.9 & 51.7 $\pm$ 3.2 & 50.8 $\pm$ 3.2 & 50.0 $\pm$ 3.2 & 50.4 $\pm$ 3.2 & 45.8 $\pm$ 4.6 & 55.0 $\pm$ 4.6 \\
\modelname-Enc-17m & 50.6 $\pm$ 1.9 & 50.0 $\pm$ 3.2 & 50.8 $\pm$ 3.2 & 50.8 $\pm$ 3.2 & 50.0 $\pm$ 3.2 & 46.7 $\pm$ 4.6 & 53.3 $\pm$ 4.6 \\
\modelname-Dec-from-Enc-17m & 49.9 $\pm$ 1.9 & 50.0 $\pm$ 3.2 & 50.0 $\pm$ 3.2 & 49.6 $\pm$ 3.2 & 49.6 $\pm$ 3.2 & 47.5 $\pm$ 4.6 & 51.7 $\pm$ 4.6 \\
\modelname-Dec-17m & 51.1 $\pm$ 1.9 & 50.0 $\pm$ 3.2 & 51.2 $\pm$ 3.2 & 52.1 $\pm$ 3.2 & 49.2 $\pm$ 3.2 & 45.0 $\pm$ 4.6 & 53.3 $\pm$ 4.6 \\
\midrule
\multicolumn{8}{c}{\textbf{XS Models (32M parameters)}} \\
\midrule
\modelname-Enc-from-Dec-32m & 53.6 $\pm$ 1.9 & 52.5 $\pm$ 3.2 & 53.8 $\pm$ 3.2 & 54.6 $\pm$ 3.2 & 53.3 $\pm$ 3.2 & 50.0 $\pm$ 4.6 & 56.7 $\pm$ 4.5 \\
\modelname-Enc-32m & 53.2 $\pm$ 1.9 & 53.8 $\pm$ 3.2 & 52.9 $\pm$ 3.2 & 52.9 $\pm$ 3.2 & 52.9 $\pm$ 3.2 & 52.5 $\pm$ 4.6 & 53.3 $\pm$ 4.6 \\
\modelname-Dec-from-Enc-32m & 51.1 $\pm$ 1.9 & 51.7 $\pm$ 3.2 & 51.2 $\pm$ 3.2 & 50.4 $\pm$ 3.2 & 51.7 $\pm$ 3.2 & 49.2 $\pm$ 4.6 & 54.2 $\pm$ 4.6 \\
\modelname-Dec-32m & 50.8 $\pm$ 1.9 & 50.4 $\pm$ 3.2 & 50.8 $\pm$ 3.2 & 51.2 $\pm$ 3.2 & 50.0 $\pm$ 3.2 & 50.0 $\pm$ 4.6 & 50.0 $\pm$ 4.6 \\
\midrule
\multicolumn{8}{c}{\textbf{Small Models (66-70M parameters)}} \\
\midrule
\modelname-Enc-from-Dec-68m & 51.9 $\pm$ 1.9 & 52.5 $\pm$ 3.2 & 51.7 $\pm$ 3.2 & 51.7 $\pm$ 3.2 & 51.2 $\pm$ 3.2 & 53.3 $\pm$ 4.6 & 49.2 $\pm$ 4.6 \\
\modelname-Enc-68m & 56.1 $\pm$ 1.9 & 55.8 $\pm$ 3.2 & 56.7 $\pm$ 3.2 & 55.8 $\pm$ 3.2 & 56.7 $\pm$ 3.2 & 56.7 $\pm$ 4.5 & 56.7 $\pm$ 4.5 \\
\modelname-Dec-from-Enc-68m & 51.8 $\pm$ 1.9 & 51.7 $\pm$ 3.2 & 52.1 $\pm$ 3.2 & 51.7 $\pm$ 3.2 & 50.8 $\pm$ 3.2 & 51.7 $\pm$ 4.6 & 50.0 $\pm$ 4.6 \\
\modelname-Dec-68m & 54.2 $\pm$ 1.9 & 55.0 $\pm$ 3.2 & 53.8 $\pm$ 3.2 & 53.8 $\pm$ 3.2 & 52.9 $\pm$ 3.2 & 56.7 $\pm$ 4.5 & 49.2 $\pm$ 4.6 \\
\midrule
\multicolumn{8}{c}{\textbf{Medium Models (150M parameters)}} \\
\midrule
\modelname-Enc-from-Dec-150m & 52.8 $\pm$ 1.9 & 50.8 $\pm$ 3.2 & 54.2 $\pm$ 3.2 & 53.3 $\pm$ 3.2 & 52.1 $\pm$ 3.2 & 47.5 $\pm$ 4.6 & 56.7 $\pm$ 4.5 \\
\modelname-Enc-150m & 57.5 $\pm$ 1.8 & 57.1 $\pm$ 3.2 & 57.9 $\pm$ 3.2 & 57.5 $\pm$ 3.2 & 57.5 $\pm$ 3.2 & 55.8 $\pm$ 4.6 & 59.2 $\pm$ 4.5 \\
\modelname--Dec-from-Enc-150m & 53.3 $\pm$ 1.9 & 52.9 $\pm$ 3.2 & 52.9 $\pm$ 3.2 & 54.2 $\pm$ 3.2 & 52.1 $\pm$ 3.2 & 55.8 $\pm$ 4.6 & 48.3 $\pm$ 4.6 \\
\modelname-Dec-150m & 54.7 $\pm$ 1.9 & 53.3 $\pm$ 3.2 & 55.8 $\pm$ 3.2 & 55.0 $\pm$ 3.2 & 52.9 $\pm$ 3.2 & 56.7 $\pm$ 4.5 & 49.2 $\pm$ 4.6 \\
\midrule
\multicolumn{8}{c}{\textbf{Large Models (400M parameters)}} \\
\midrule
\modelname-Enc-from-Dec-400m & 55.1 $\pm$ 1.9 & 55.4 $\pm$ 3.2 & 54.6 $\pm$ 3.2 & 55.4 $\pm$ 3.2 & 55.4 $\pm$ 3.2 & 53.3 $\pm$ 4.6 & 57.5 $\pm$ 4.5 \\
\modelname-Enc-400m & 70.3 $\pm$ 1.7 & 68.8 $\pm$ 3.0 & 70.8 $\pm$ 2.9 & 71.2 $\pm$ 2.9 & 69.2 $\pm$ 3.0 & 69.2 $\pm$ 4.2 & 69.2 $\pm$ 4.2 \\
\modelname-Dec-from-Enc-400m & 54.0 $\pm$ 1.9 & 53.8 $\pm$ 3.2 & 55.0 $\pm$ 3.2 & 53.3 $\pm$ 3.2 & 52.5 $\pm$ 3.2 & 55.8 $\pm$ 4.6 & 49.2 $\pm$ 4.6 \\
\modelname-Dec-400m & 55.3 $\pm$ 1.9 & 54.6 $\pm$ 3.2 & 55.8 $\pm$ 3.2 & 55.4 $\pm$ 3.2 & 52.9 $\pm$ 3.2 & 51.7 $\pm$ 4.6 & 54.2 $\pm$ 4.6 \\
\midrule
\multicolumn{8}{c}{\textbf{XL Models (1B parameters)}} \\
\midrule
\modelname-Enc-from-Dec-1B & 57.9 $\pm$ 1.8 & 56.7 $\pm$ 3.2 & 58.3 $\pm$ 3.2 & 58.8 $\pm$ 3.2 & 54.2 $\pm$ 3.2 & 48.3 $\pm$ 4.6 & 60.0 $\pm$ 4.5 \\
\modelname-Enc-1B & 68.2 $\pm$ 1.7 & 67.1 $\pm$ 3.0 & 66.2 $\pm$ 3.1 & 71.2 $\pm$ 2.9 & 65.8 $\pm$ 3.1 & 65.8 $\pm$ 4.3 & 65.8 $\pm$ 4.3 \\
\modelname-Dec-from-Enc-1B & 55.8 $\pm$ 1.9 & 55.8 $\pm$ 3.2 & 56.7 $\pm$ 3.2 & 55.0 $\pm$ 3.2 & 54.2 $\pm$ 3.2 & 54.2 $\pm$ 4.6 & 54.2 $\pm$ 4.6 \\
\modelname-Dec-1B & 56.7 $\pm$ 1.8 & 56.7 $\pm$ 3.2 & 55.4 $\pm$ 3.2 & 57.9 $\pm$ 3.2 & 52.1 $\pm$ 3.2 & 50.8 $\pm$ 4.6 & 53.3 $\pm$ 4.6 \\
\bottomrule
\end{tabular}
\caption{WinoGender accuracy results (values: Accuracy \% $\pm$ Std Error \%). Results taken from the Eleuther AI harness. Many of the small models do not get above random performance (50\%).}
\end{table*}

\begin{figure*}[t!]
  \centering
\includegraphics[width=\textwidth]{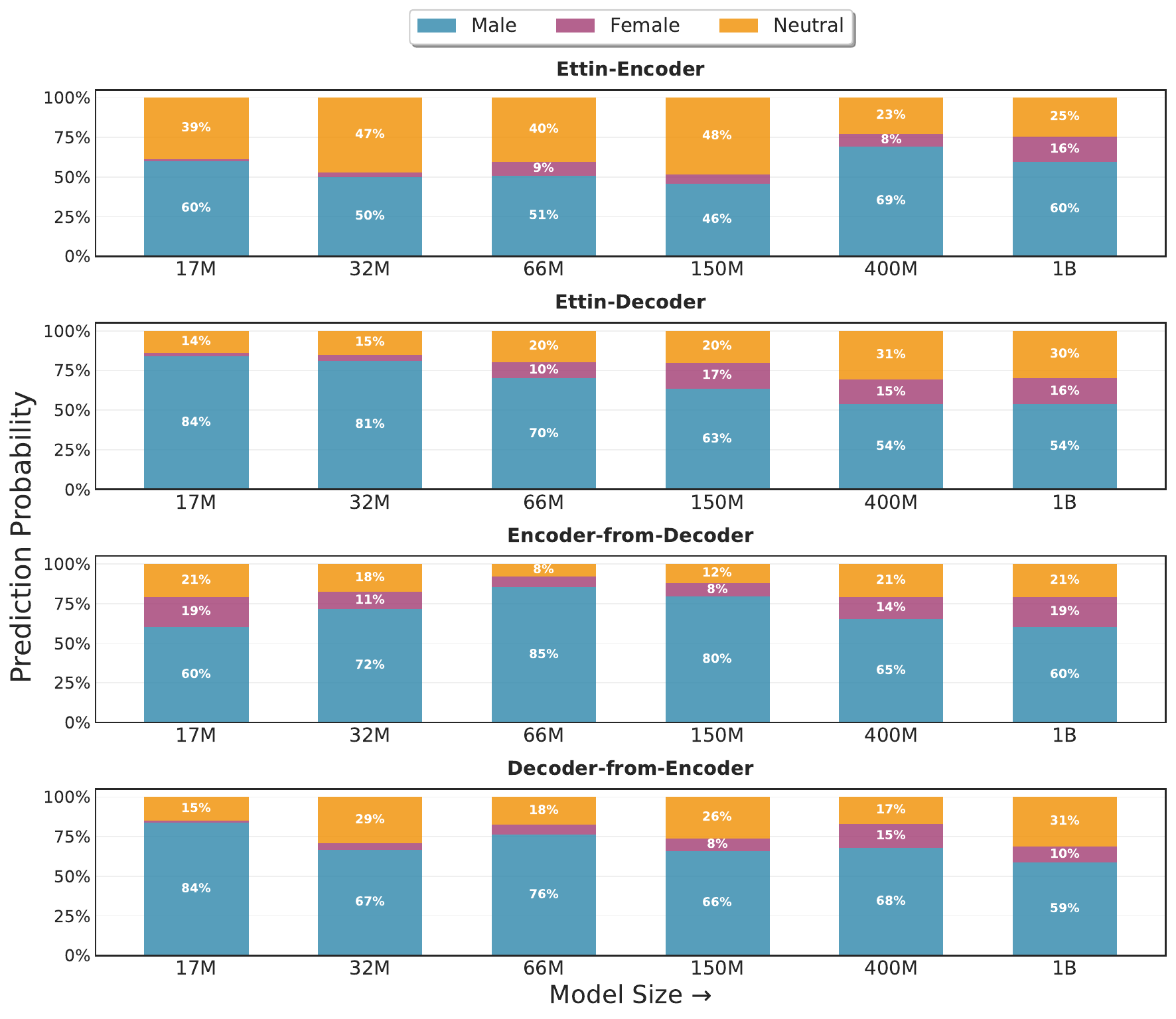}
  \caption{Full gender pronoun predictions results on the Gotcha split of WinoGender \citep{rudinger2018gender}, a 50/50 stereotypical split. \textbf{We see that encoder models are more likely to use gender neutral pronouns whereas both are biased towards male pronouns.}}
  \label{fig:bias_all}
 \vspace{-0.5em}
\end{figure*}

\begin{table*}[h!]
\centering
\small
\begin{tabular}{lr|rr}
\toprule
\textbf{Model Name} & \textbf{Total Params} & \textbf{Embed Params} & \textbf{Non-Embed Params} \\
\midrule
\multicolumn{4}{c}{\textbf{XXS Models (7-17M parameters)}} \\
\midrule
Pythia-14m & 7.6M & 6.4M & 1.2M \\
BERT Tiny & 11.2M & 7.9M & 3.2M \\
TinyBERT & 14.4M & 9.7M & 4.7M \\
\modelname-17m & 16.8M & 12.9M & 3.9M \\
\midrule
\multicolumn{4}{c}{\textbf{XS Models (28-33M parameters)}} \\
\midrule
BERT Small & 28.8M & 15.9M & 12.9M \\
\modelname-32m & 31.9M & 19.3M & 12.5M \\
MiniLM L12 & 33.4M & 11.9M & 21.4M \\
\midrule
\multicolumn{4}{c}{\textbf{Small Models (68-82M parameters)}} \\
\midrule
DistilBERT Base & 66.4M & 23.8M & 42.5M \\
\modelname-68m & 68.1M & 25.8M & 42.4M \\
DistilGPT2 & 81.9M & 39.4M & 42.5M \\
DistilRoBERTa Base & 82.1M & 39.0M & 43.1M \\
\midrule
\multicolumn{4}{c}{\textbf{Base Models (123-150M parameters)}} \\
\midrule
BERT-base & 109.5M & 23.8M & 85.6M \\
Pythia-160m & 123.7M & 38.6M & 85.1M \\
SmolLM2-135m & 134.5M & 28.3M & 106.2M \\
ModernBERT-base & 149.0M & 38.7M & 110.3M \\
\modelname-150m & 149.0M & 38.7M & 110.3M \\
\midrule
\multicolumn{4}{c}{\textbf{Large Models (353-395M parameters)}} \\
\midrule
BERT-large & 335.1M & 31.8M & 303.4M \\
Pythia-410m & 353.8M & 51.5M & 302.3M \\
SmolLM2-360m & 361.8M & 47.2M & 314.6M \\
ModernBERT-large & 394.8M & 51.6M & 343.2M \\
\modelname-400m & 394.8M & 51.6M & 343.2M \\
\midrule
\multicolumn{4}{c}{\textbf{XL Models (884M-1.2B parameters)}} \\
\midrule
DeBERTa v2 XLarge & 884.6M & 197.6M & 687.0M \\
Pythia 1B & 908.8M & 103.0M & 805.7M \\
\modelname-1B & 1028.1M & 90.3M & 937.8M \\
OLMo 1B 0724 & 1176.8M & 103.0M & 1073.7M \\
Llama 3.2 1B & 1235.8M & 262.7M & 973.1M \\
\bottomrule
\end{tabular}
\caption{Parameter breakdown of language models organized by size categories. Models are grouped by total parameter count and show the distribution between embedding and non-embedding parameters across different architectures. Parameter counts are the same for \modelname\ encoders, decoders, and cross-objective trained versions.}
\label{tab:model_params}
\end{table*}

\begin{table}[htbp]
\centering
\label{tab:common}
\begin{tabular}{lr}
\toprule
\textbf{Parameter} & \textbf{Value} \\
\midrule
Vocabulary Size & 50,368 \\
Max Sequence Length & 1024->7999 \\
Tokenizer & ModernBERT \\
Attention Layer & RoPE \\
Attention Dropout & 0.0 \\
Attention Output Bias & false \\
Attention Output Dropout & 0.1 \\
Attention QKV Bias & false \\
Transformer Layer & prenorm \\
Embedding Dropout & 0.0 \\
Embedding Norm & true \\
Final Norm & true \\
Skip First PreNorm & true \\
Embedding Layer & sans\_pos \\
MLP Dropout & 0.0 \\
MLP Input Bias & false \\
MLP Layer Type & GLU \\
MLP Output Bias & false \\
Normalization & LayerNorm \\
Norm Epsilon & 1e-12 \\
Norm Bias & false \\
Hidden Activation & GELU \\
Head Pred Activation & GELU \\
Activation Function & GELU \\
Padding & unpadded \\
Rotary Embedding Base & 160,000.0 \\
Rotary Embedding Interleaved & false \\
Allow Embedding Resizing & true \\
Sliding Window & 128 \\
Global Attention Every N Layers & 3 \\
Unpad Embeddings & true \\
\bottomrule
\end{tabular}
\vspace{0.75em}
\caption{Common pre-training configuration parameters across all models}
\end{table}

\end{document}